\documentclass[transmag]{IEEEtran}
\usepackage{latexsym}
\usepackage{graphicx}
\usepackage{amsfonts,amssymb,amsmath}
\usepackage{hyperref}
\usepackage{xcolor}
\usepackage{subcaption}
\usepackage{booktabs} % For formal tables
\usepackage[linesnumbered,ruled,vlined]{algorithm2e} % For algorithms 
\usepackage{algpseudocode} 
\usepackage{graphicx}
\usepackage{subcaption}
\usepackage{tabularx}
\usepackage{epsfig}  
\usepackage{xcolor}
\usepackage{amsthm}
\usepackage{multirow}
\usepackage{amssymb}
\usepackage{xcolor}
\usepackage{pdfpages}
\usepackage{url}
\usepackage{fancyhdr}
\usepackage{soul}
\usepackage{bbm}
\usepackage{enumitem}

\graphicspath{{./figs/}{./plots}{./images/}}

\newcommand{\best}[1]{\color{red}{#1}}
\newcommand{\secondbest}[1]{\color{blue}{#1}}

\begin{document}

\title{Fleet Rebalancing for Expanding Shared e-Mobility Systems: \\ 
A Multi-agent Deep Reinforcement Learning Approach}

\author{%
  Man Luo\textsuperscript{1,2},  Bowen Du\textsuperscript{3}, Wenzhe Zhang\textsuperscript{$\dagger$4}, Tianyou Song\textsuperscript{$\dagger$5}, Kun Li\textsuperscript{$\dagger$6}\thanks{$\dagger$ Work done during an internship at the University of Warwick.},   
  Hongming Zhu\textsuperscript{7}, \\ Mark Birkin\textsuperscript{2,8}, Hongkai Wen\textsuperscript{*2,3} 
  \\
  \textsuperscript{1} University of Exeter, UK \hspace*{10pt}
  \textsuperscript{2} The Alan Turing Institute, UK \hspace*{10pt} 
  \textsuperscript{3} University of Warwick, UK \\
  \textsuperscript{4} University of California San Diego, US \hspace*{10pt}
  \textsuperscript{5} University of Illinois Urbana-Champaign, US  \\ \textsuperscript{6} Columbia University, US  \hspace*{10pt}
   \textsuperscript{7} Tongji University, China \hspace*{10pt}
   \textsuperscript{8} University of Leeds, UK
   
 \\

}

%%%%%%%%%%%%%%%%%%%%%%%%%%%%%%%%%%%%%%%%%%%%%%%%%%%
%% Abstract
%%%%%%%%%%%%%%%%%%%%%%%%%%%%%%%%%%%%%%%%%%%%%%%%%%%

\IEEEtitleabstractindextext{

\begin{abstract}
The electrification of shared mobility has become popular across the globe. Many cities have their new shared e-mobility systems deployed, with continuously expanding coverage from central areas to the city edges. A key challenge in the operation of these systems is fleet rebalancing, i.e., how EVs should be repositioned to better satisfy future demand. This is particularly challenging in the context of expanding systems, because i) the range of the EVs is limited while charging time is typically long, which constrain the viable rebalancing operations; and ii) the EV stations in the system are dynamically changing, i.e., the legitimate targets for rebalancing operations can vary over time. We tackle these challenges by first investigating rich sets of data collected from a real-world shared e-mobility system for one year, analyzing the operation model, usage patterns and expansion dynamics of this new mobility mode. With the learned knowledge we design a high-fidelity simulator, which is able to abstract key operation details of EV sharing at fine granularity. Then we model the rebalancing task for shared e-mobility systems under continuous expansion as a Multi-Agent Reinforcement Learning (MARL) problem, which directly takes the range and charging properties of the EVs into account. We further propose a novel policy optimization approach with action cascading, which is able to cope with the expansion dynamics and solve the formulated MARL. We evaluate the proposed approach extensively, and experimental results show that our approach outperforms the state-of-the-art, offering significant performance gain in both satisfied demand and net revenue.

\end{abstract}


}
\maketitle

%%%%%%%%%%%%%%%%%%%%%%%%%%%%%%%%%%%%%%%%%%%%%%%%%%%
%% Intro
%%%%%%%%%%%%%%%%%%%%%%%%%%%%%%%%%%%%%%%%%%%%%%%%%%%

\section{Introduction}
A promising trend of future mobility is electric and shared. Recently, EV sharing systems have been expanding fast in major cities around the world, from London~\cite{Web:Bluecity} to Berlin~\cite{Web:Weshare} and Singapore~\cite{Web:Bluesg}. They provide a convenient way for users to pick up a shared Electric Vehicle (EVs) from nearby stations and drive around the city whenever needed, offering a more sustainable and efficient mobility option to all citizens, independent of individual car ownership.

\begin{figure}[t]
\centering
\includegraphics[width=\columnwidth]{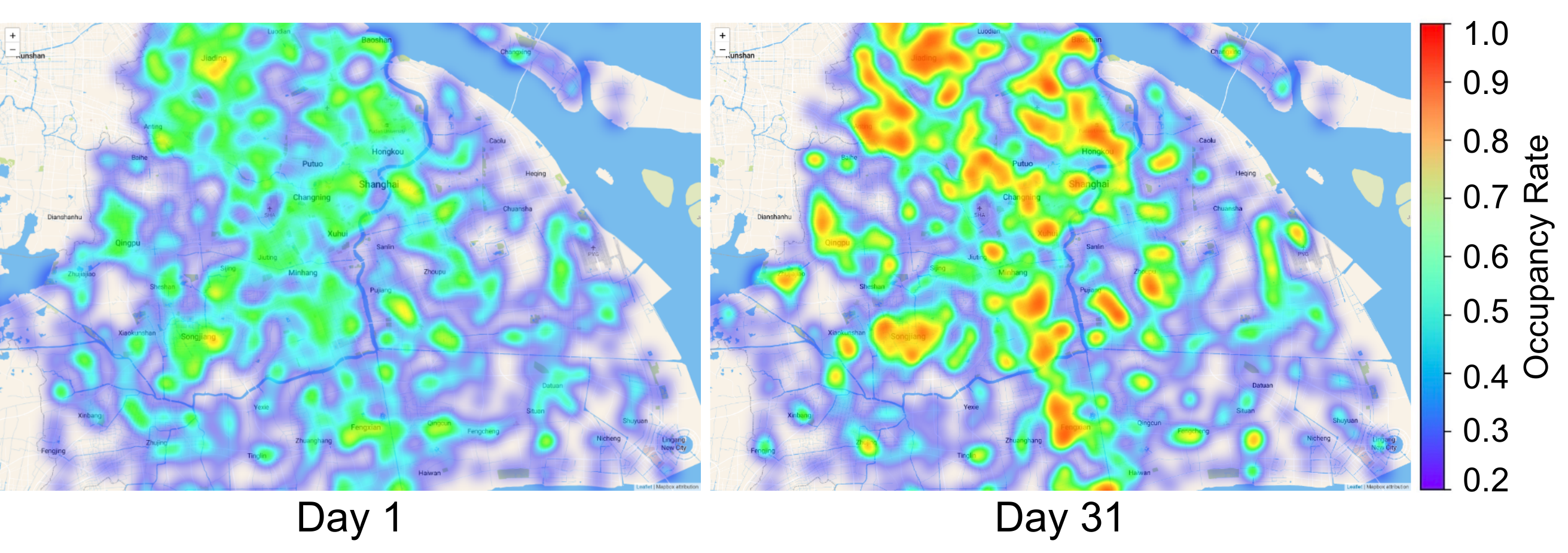}
\caption{Distribution of occupancy rate (the ratio of parked vehicles to the total number of available parking spaces) across stations of a shared e-mobility system over a month.} 
\vspace{-2mm}
\label{fig:imbalance}
\end{figure}

Despite their rapid adoption across the globe, a major challenge in the operation of shared e-mobility systems is the imbalanced fleet distribution over time. For instance, Fig.~{\ref{fig:imbalance}} shows the spatial distribution of station occupancy rate (ratio of parked vehicles to the total available space) in a real-world shared e-mobility system at the beginning (Day 1, left) and the end (Day 31, right) of a month respectively. Note that here at Day 1 the system has just been manually rebalanced. We see that after 30 days the distribution becomes very skewed, where stations in some areas (shown by the red patches) have substantially more EVs accumulated than the others. In addition, such imbalance also happens within shorter time frames, e.g., we observe that in morning rush hours a large volume of EVs tend to flow to central areas and stay there, making fewer or even no vehicle available in other places. This certainly poses negative impact on the overall system performance, as customers may refrain from using the system if there is no EVs available nearby, or no free parking spaces at their destinations.

In fact this is a common issue in many types of shared mobility systems, e.g. shared bikes~\mbox{\cite{ghosh:AIJ:2017,li:Kdd:2018,ghosh:IJCAI:2019}}, taxi~\mbox{\cite{lin:Kdd:2018,Li:WWW:2019,wei:IEEEACESS:2017look}}, and ride-sharing services~\mbox{\cite{kooti:WWW:2017,jiang:www:2018}}, and the de facto solution is to rebalance the fleet during operation. In practice, there are mainly two classes of rebalancing strategies: i) employing dedicated teams to manually reposition the vehicles/bikes across space~{\cite{li:Kdd:2018}}; or ii) incentivizing the users or drivers to voluntarily rent/return vehicles/bikes to the desired locations~\mbox{\cite{singla:AAAI:2015,pan:AAAI:2019}}. In this paper, we adopt the latter, i.e. user incentive based rebalancing, since it is natural in our context to offer monetary incentives e.g., price discount to the users in exchange for them to reposition the EVs to alternative destinations. However, unlike the other mobility systems, the rebalancing problem in shared e-mobility systems has its unique challenges. First of all, EVs typically have limited range, while charging takes much longer than filling up traditional vehicles. This imposes many implicit constraints on rebalancing strategies, e.g. EVs can't be repositioned to locations that are beyond the remaining range, which also need to be sufficiently charged to serve any future orders.
Secondly, as EV sharing is relatively new to many cities, at this stage they are still expanding their infrastructure continuously: for instance, the system studied in this paper almost doubled its stations during one year, with hundreds of new stations being deployed in each month. This makes the rebalancing task even more challenging, as each time the candidate stations to which the EVs may be repositioned are dynamically changing.

In this paper, we tackle the rebalancing problem in the expanding shared e-mobility systems with a novel multi-agent reinforcement learning (MARL) approach. We model the rebalancing task as a stochastic game among multiple agents, each of which manages the EV stations within a spatial region, and learns to make informed decisions as where to reposition the incoming EVs by incentivizing the users. To address the challenges of limited EV range and charging delays, we propose to incorporate the range and charging information directly in our MARL algorithm, so that the agents are fully aware of those constraints when making decisions. To cope with the dynamically expanding station infrastructure, we design a novel action cascading approach, which decomposes the actions of repositioning an EV into two subsequent and conditionally dependent sub-actions. The intuition is that when an EV needs to be repositioned, one could firstly decide which of the regions it should be redirected to, and then given the decision subsequently determine which station within that region should be the new destination. Therefore, the expansion dynamics are localized within individual regions, while the first sub-actions should have static action spaces. In light of this, the proposed action cascading approach uses two connected policy networks to generate the sub-actions in sequel, where the second network is specially designed to handle the non-stationary action spaces.

There is also a solid body of existing work that adopts MARL formulation in rebalancing shared mobility systems. For instance, the recent work in{~\cite{li:Kdd:2018}} uses a spatial-temporal deep Q networks (DQN) to rebalance shared bikes, but it is fundamentally different from our work since it does not consider the dynamic system expansion. On the other hand, the work in~{\cite{Li:WWW:2019}} and~{\cite{lin:Kdd:2018}} study the order dispatching problem in taxi systems. Although different from our problem, these work share similar challenges as they also consider changing action spaces. However, their solutions are to allow the agents to directly rank the potential actions and select the one with the highest score, while we use two policy networks to generate cascading actions and handle the non-stationarity. Concretely, the contributions of this paper are as follows: 

\begin{itemize}[leftmargin=16pt, topsep=0pt]
    \item To the best of our knowledge, we are the first to identify the problem of rebalancing the continuously expanding EV sharing systems, and formulate the incentive-based rebalancing problem under the multi-agent reinforcement learning framework. 
    
    \item We conduct an in-depth case study of a real-world shared e-mobility system for one year, collected and analyzed multiple modalities of data, which provides key insights on the operation models, expansion processes and usage patterns of the shared e-mobility systems.
    
    \item We propose a novel policy optimization approach with action cascading (ac-PPO), which uses two connected policy networks to handle the dynamics introduced by the continuous expansion of the shared e-mobility systems. 
    
    \item We design and open-source a high-fidelity simulator\footnote{Available at https://github.com/manluow/ev-simulator} to simulate both the operation and expansion of shared e-mobility systems, which is calibrated with real-world data to support training and evaluation of the proposed reinforcement learning approaches.
    
    \item The proposed approach has been evaluated extensively, and the results show that our approach significantly outperforms the state of the art, offering up to 12\% improvement in net revenue and 14\% in demand satisfied rate. 
    
\end{itemize}

%%%%%%%%%%%%%%%%%%%%%%%%%%%%%%%%%%%%%%%%%%%%%%%%%%%
%% EV
%%%%%%%%%%%%%%%%%%%%%%%%%%%%%%%%%%%%%%%%%%%%%%%%%%%

\section{Shared e-Mobility Systems in the Real World}
\label{sec:ev-sharing}

\subsection{The EV Sharing Model}
\label{sub:ev-model}
% Electric Vehicles (EV)
\noindent \textbf{Electric Vehicles (EVs): }
We assume that in the EV sharing context, the vehicles are of limited range, depending on the particular EV models used in the system. Without loss of generality, we assume the shared e-mobility system has one EV model in operation, with a fixed range and a typical discharging model~{\cite{Tremblay:Charging:2009}}. The charging time is governed by a charging model~{\cite{Tremblay:Charging:2009}}, the remaining range, battery capacities and charger specifications.

% EV Sharing Stations
\noindent \textbf{EV Sharing Stations: }
The shared e-mobility systems considered in this paper are \emph{station based}, i.e. the users only pick up or return EVs from/to the available stations. Concretely, we represent a station $s$ as a tuple $(\texttt{loc}, \#\texttt{c}, \#\texttt{v})$, where \texttt{loc} is the station coordinates (e.g. latitude and longitude), \#\texttt{c} is the total number of charging docks (i.e. parking spaces) within the station, and \#\texttt{v} is the number of EVs initially equipped in the station. Typically we assume $\#\texttt{v}<\#\texttt{c}$, i.e. when station $s$ was newly deployed, it had \#\texttt{v} EVs available for pick up and \#\texttt{c} - \#\texttt{v} free spaces for returns. 

% User Orders 
\noindent \textbf{Shared e-Mobility Systems: }
Given the set of available stations $S$, the shared e-mobility system operates as follows. Assume at time $t$, a user would like to rent an EV from a station $s^{\text{o}}\in S$, and return to another $s^{\text{d}}\in S$. If she finds at least one EV available at the picking up station $s^{\text{o}}$, that has been sufficiently charged with enough range to cover the planned trip, she will post an order $o_t = (s^{\text{o}}, s^{\text{d}})$ to the shared e-mobility system, which would allow her to pick up the EV and log the start of this order. The order is considered to be completed when the EV is returned to the destination station $s^{\text{d}}$ and plugged in for charging, whose price is calculated based on rental duration. For simplicity, we assume that all orders accepted by our system will be completed, i.e., there is no change or cancellation during the trips.

% System expansion
\noindent \textbf{Dynamic System Expansion: }
In this paper, we consider dynamic expansion of the shared e-mobility system, i.e., during operation the system have new stations deployed while existing ones closed at arbitrary time. This is common in the real world, as new stations could be deployed to extend system coverage in new areas, or within the already covered areas to increase station density. On the other hand, stations can also be closed temporarily or permanently, e.g., where/when profit becomes limited or parking spaces no longer available. We assume that overall the system keeps expanding, i.e., typically more stations are deployed than closed.

 \begin{figure*}[ht]
 \centering
 \begin{subfigure}{0.297\textwidth}
 \includegraphics[width=\textwidth]{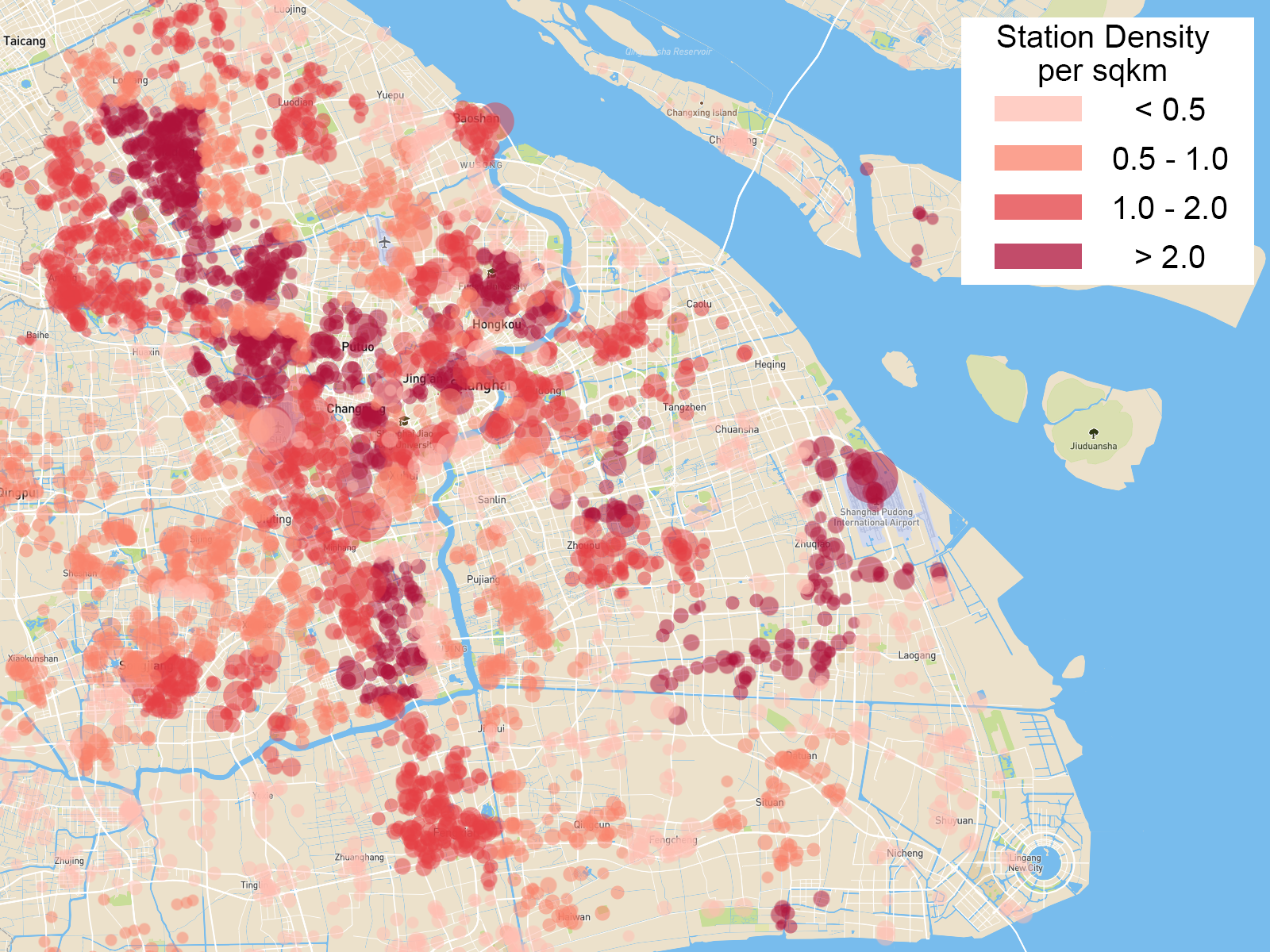}
 \caption{}
 \label{fig:spatial-stations}
 \end{subfigure}\hspace{3mm}
  \begin{subfigure}{0.3\textwidth}
 \includegraphics[width=\textwidth]{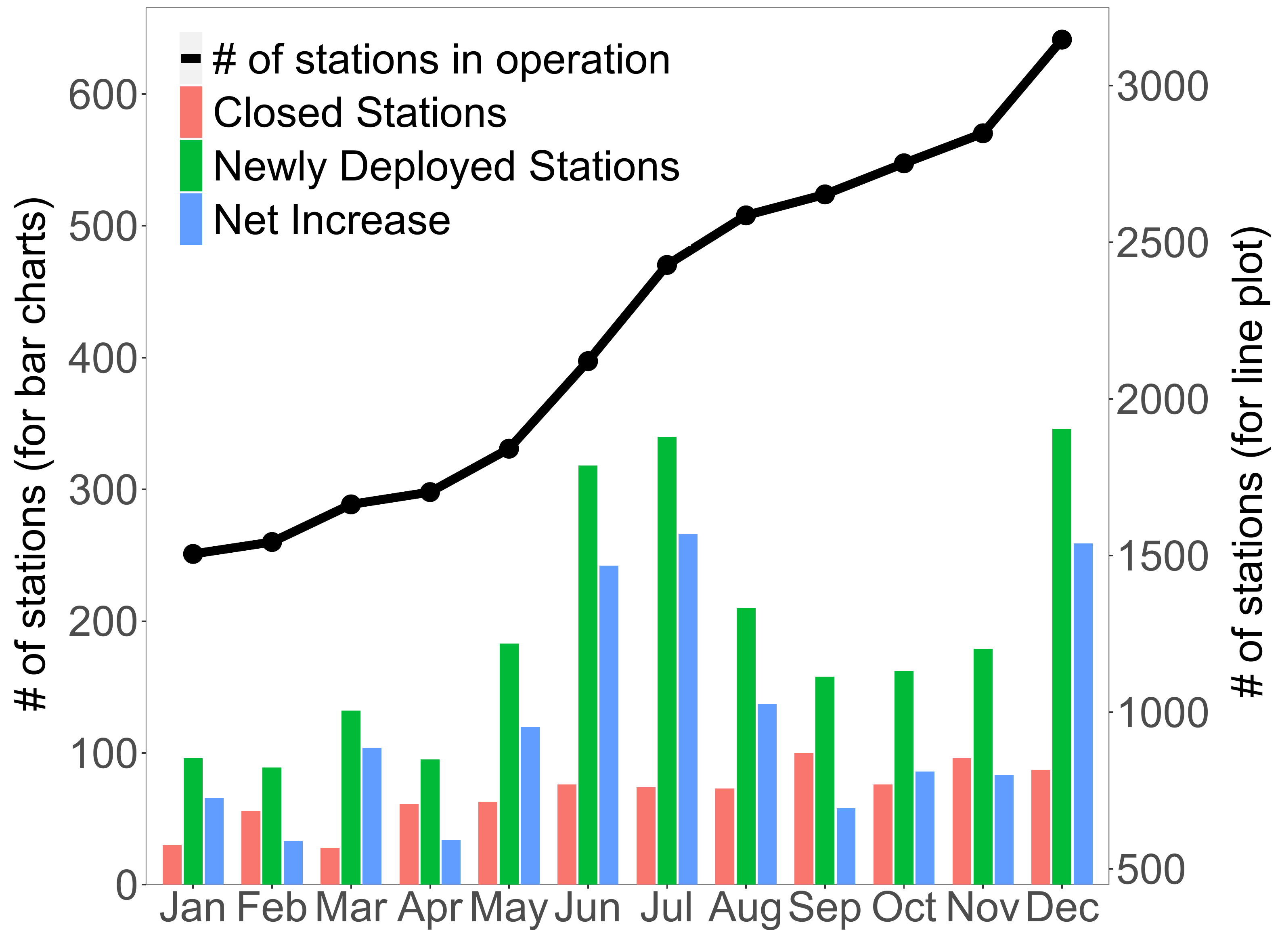}
 \caption{}
  \label{fig:expansion-stats}
 \end{subfigure}\hspace{3mm}
 \begin{subfigure}{0.33\textwidth}
 \includegraphics[width=\textwidth]{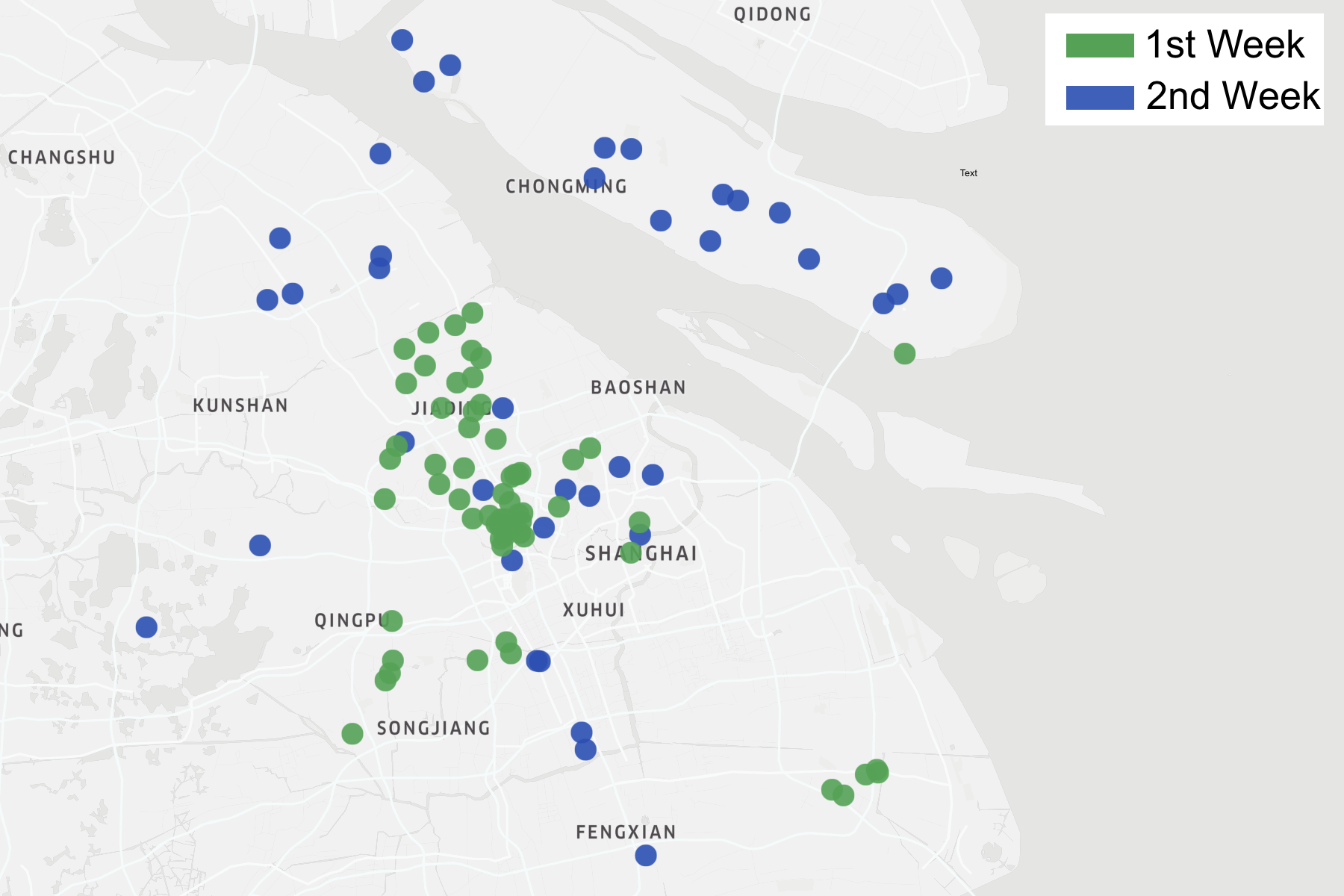}
 \caption{}
 \label{fig:spatial-deploy}
 \end{subfigure}

 \caption{(a) The distribution, density and sizes of the current stations in the system. (b) Statistics of system expansion during 12 months period (left $y$-axis for bar charts, right for line plot). (c) Newly deployed stations in two consecutive weeks.}
 \label{fig:spatial}
 \end{figure*}

\begin{figure*}[t]
 \centering
 \begin{subfigure}{0.36\textwidth}
 \includegraphics[width=\textwidth]{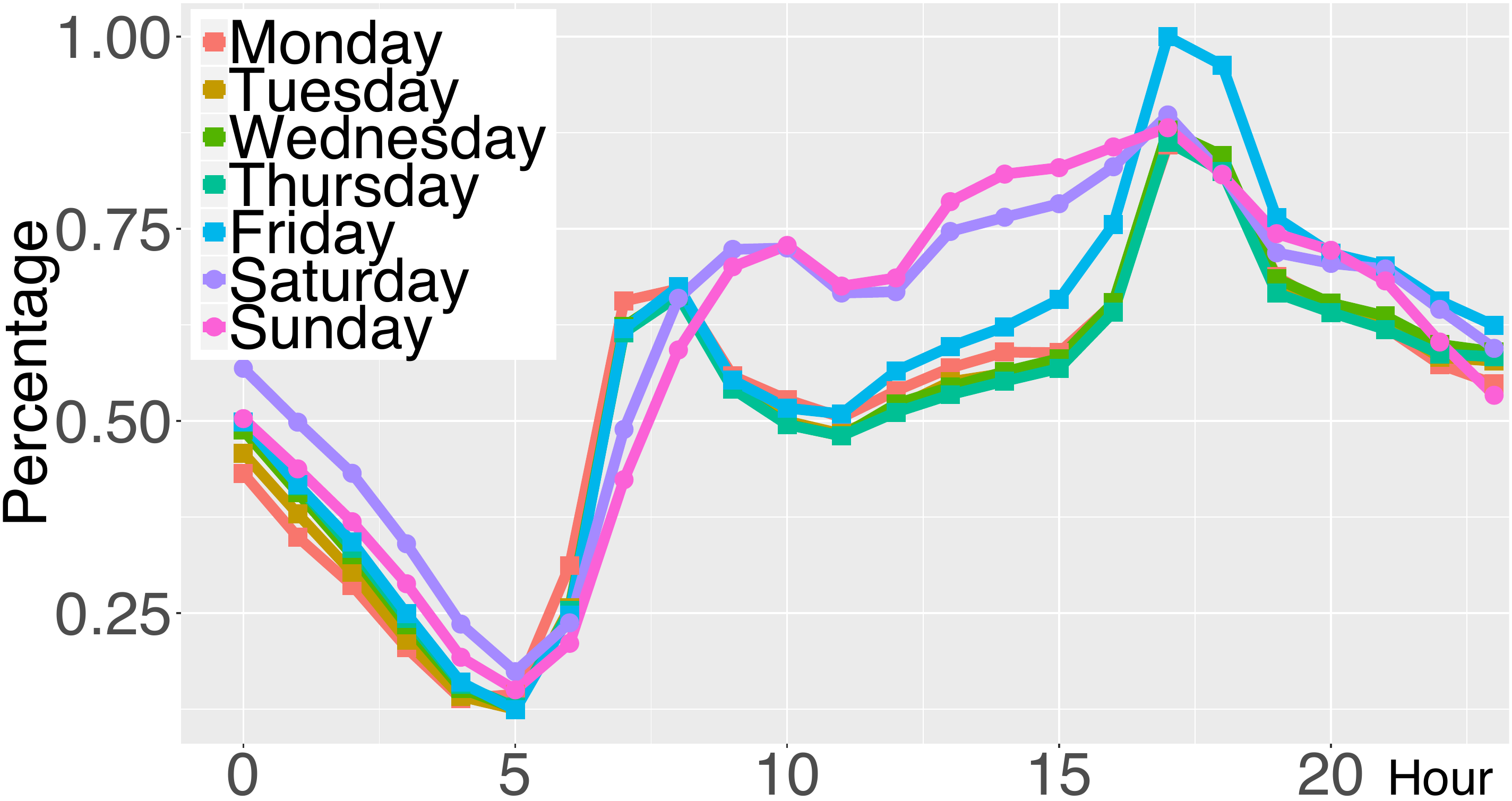}
 \caption{}
 \label{fig:week}
 \end{subfigure} \hspace{5mm}
 \begin{subfigure}{0.25\textwidth}
 \includegraphics[width=\textwidth]{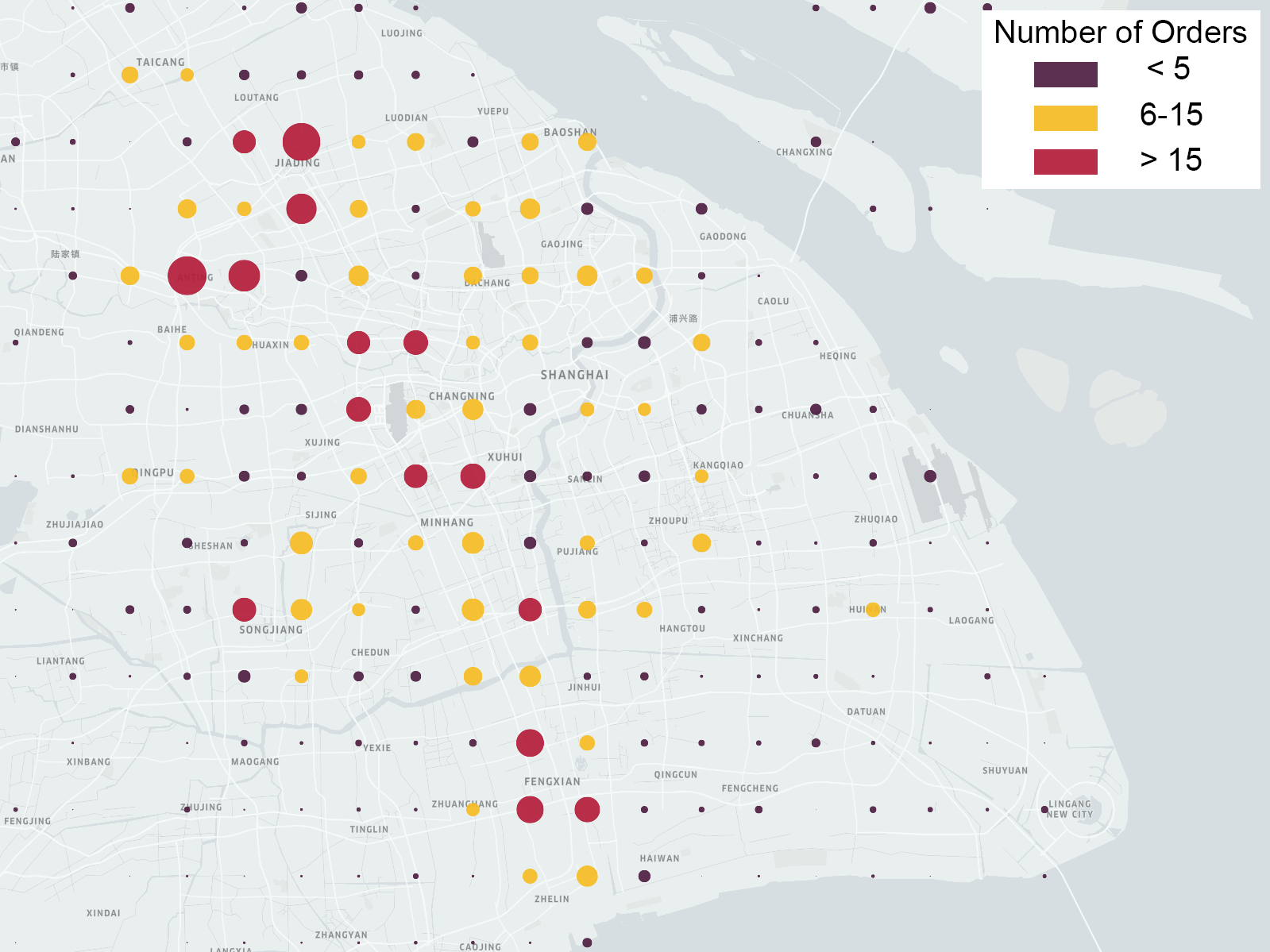}
 \caption{}
 \label{fig:patterns-morning}
 \end{subfigure} \hspace{5mm}
 \begin{subfigure}{0.25\textwidth}
 \includegraphics[width=\textwidth]{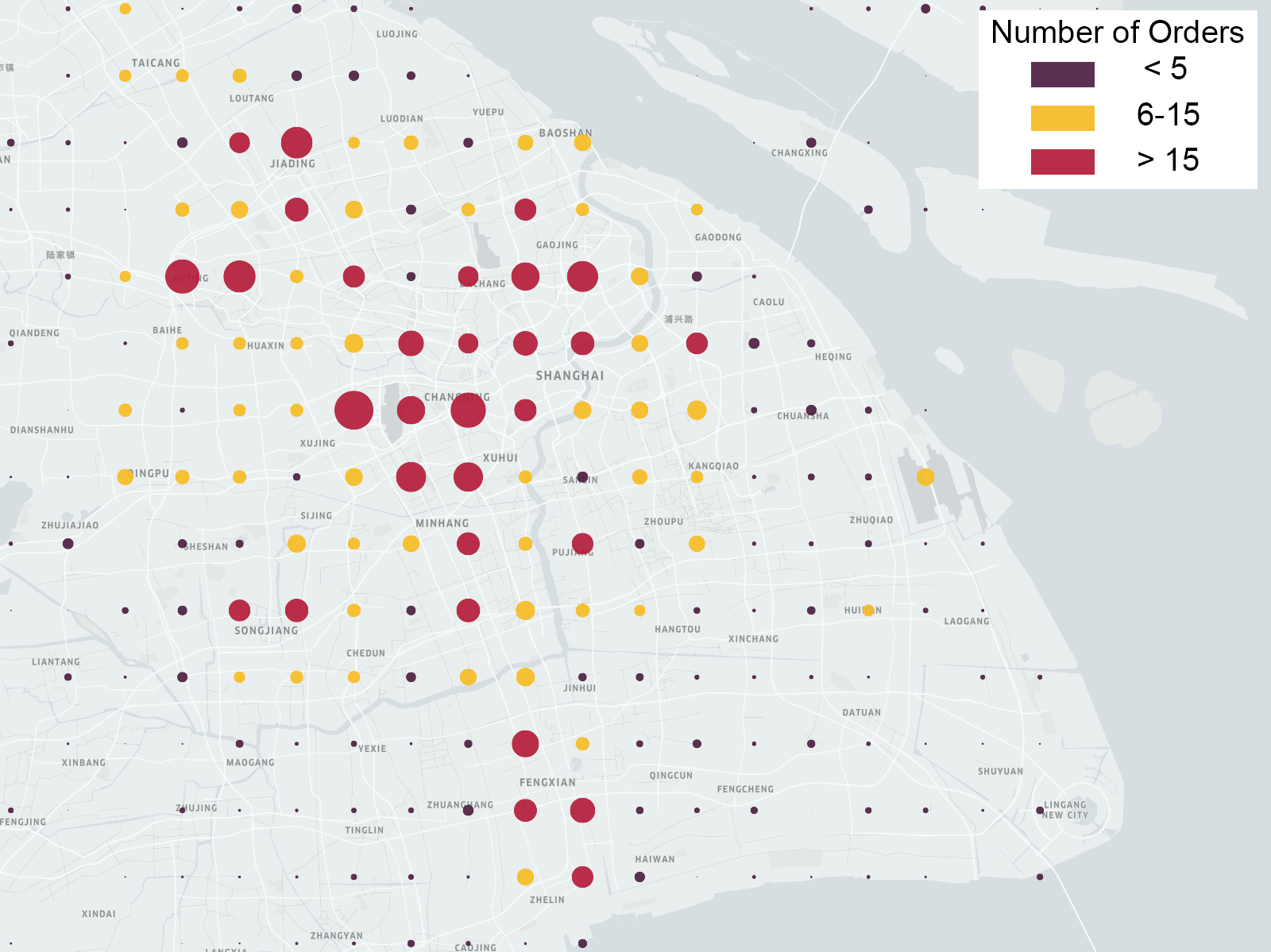}
 \caption{}
 \label{fig:patterns-evening} 
 \end{subfigure} 
  \caption{Demand patterns can be highly imbalanced across space and time. Average number of orders during (a) each day in a week, (b) morning rush hours, and (c) evening rush hours.}
 \label{fig:patterns}
 \vspace{-5mm}
 \end{figure*}

\subsection{Analysis of the EV Sharing Data}
\label{sub:ev-data}
% \noindent \textbf{Data Collected. }
For our study, we worked with a major shared e-mobility provider in Shanghai and collected its operational data for one year, including i) the complete \textit{order transactions} of the system, and ii) records on \textit{station deployment}, i.e., when and where a station was deployed or closed. The order data contains detailed information about each transaction, including the anonymous user ID who initiated the order, the origin/destination stations, the timestamps of pick-up/return, the total duration of the order and the final price etc. In total we have over 7 million valid order transactions, which were generated by approximately 0.4 million active users during the one year period. For the station deployment data, we first collected the system status at the beginning of the year, which had approximately 1,700 stations and 4,700 EVs in operation. We also logged the system expansion process, i.e. when and where new stations were deployed, and by the end of the year, the system had more than 4,000 stations and 17,000 EVs. 

In the following, we discuss key insights obtained from the collected data, from characteristics of infrastructure deployment and expansion, to its usage patterns and user behaviours in the spatio-temporal domain. This not only sheds light on how rebalancing should be carried out appropriately, but can also be generalized to and benefit other tasks, such as dynamic pricing and station optimization in shared mobility.
 
\noindent \textbf{Station Distribution. }  
Fig.~\ref{fig:spatial-stations} shows the distribution, density and sizes (\#\texttt{c}) of the current stations in shared e-mobility system. Firstly, it is clear that the density of the deployed stations is very different across the space. For instance, we find that the density decreases significantly towards the outer ring road, e.g., from 3.23 stations per $km^2$ in the inner ring road, to only 0.93 stations per $km^2$ in areas beyond the outer ring road. This would certainly affect the rebalancing decisions as in denser areas there may be more choices for alternative destinations, which are also closer. We also see that across the city the sizes of stations vary, and there are some particularly large stations. By cross-checking the nearby POIs we find that those stations are typically located in transportation hubs such as airports, which are key for the rebalancing task, as there should be constantly significant volume of in/out demand, and need sufficient amount of available EVs.

\noindent \textbf{System Expansion. }  
Fig.~\ref{fig:expansion-stats} visualizes the expansion of shared e-mobility system studied in this paper. We see that in 12 months time, the stations in operation has doubled from roughly 1,500 to more than 3,000, where in each month there are continuously hundreds of stations being deployed or closed. This would pose significant challenges to rebalancing task, as the possible destinations for repositioning the EVs are continuously changing. Another observation is that the expansion process is not uniform, which requires the rebalancing strategies to adapt. For instance, Fig.~{\ref{fig:spatial-deploy}} shows the newly deployed stations of the service in two consecutive weeks. We see that during the first week more stations were deployed at the central areas with only a few scattered around, while in latter week stations were spread more uniformly. This indicates that as the system expands over time, different regions across the city might require different rebalancing strategies to better exploit the available infrastructure.

 \begin{figure*}[t]
 \centering
  \begin{subfigure}{0.61\columnwidth}
 \includegraphics[width=\textwidth]{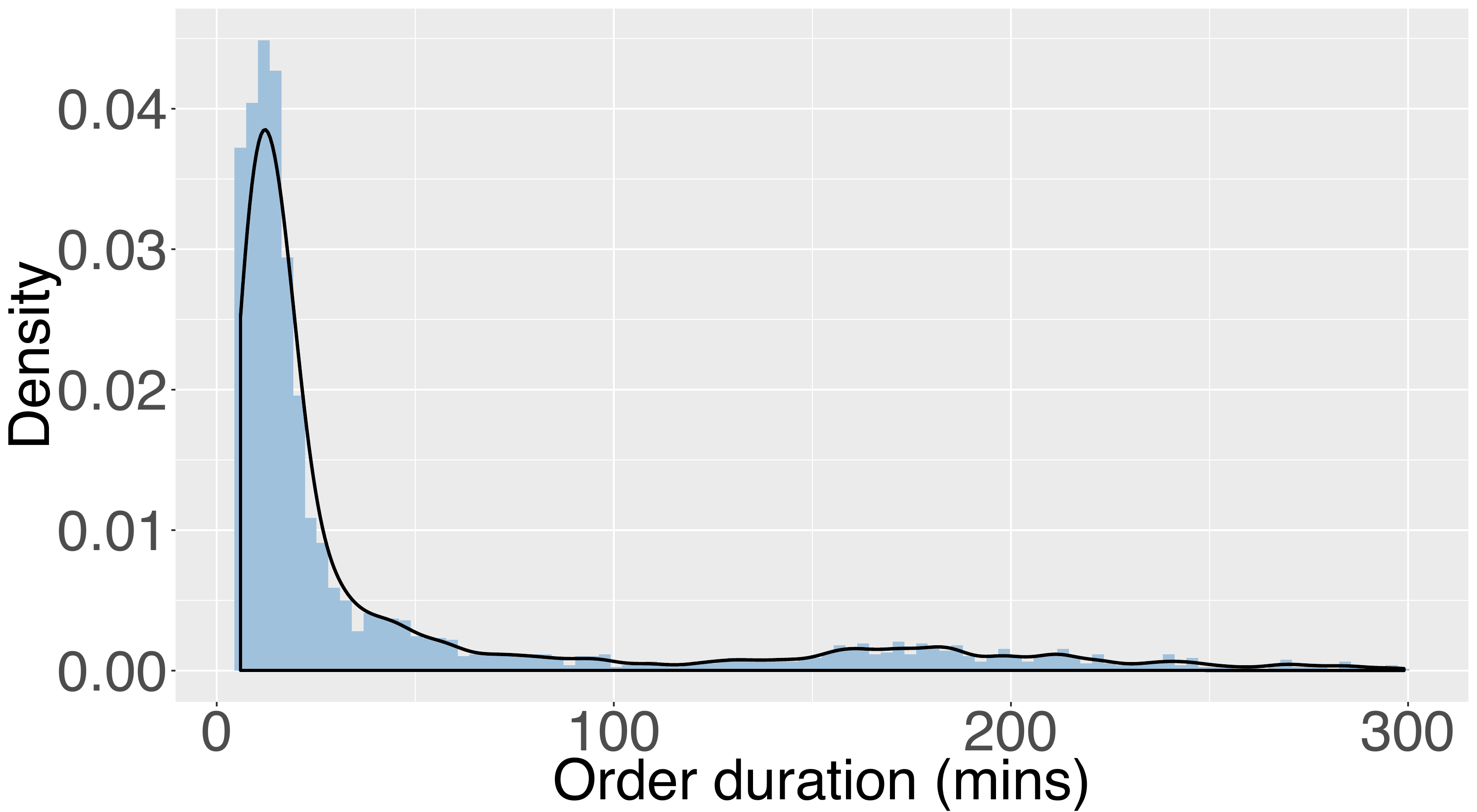}
  \caption{}
 \label{fig:order-len} 
 \end{subfigure} 
 \begin{subfigure}{0.61\columnwidth}
 \includegraphics[width=\textwidth]{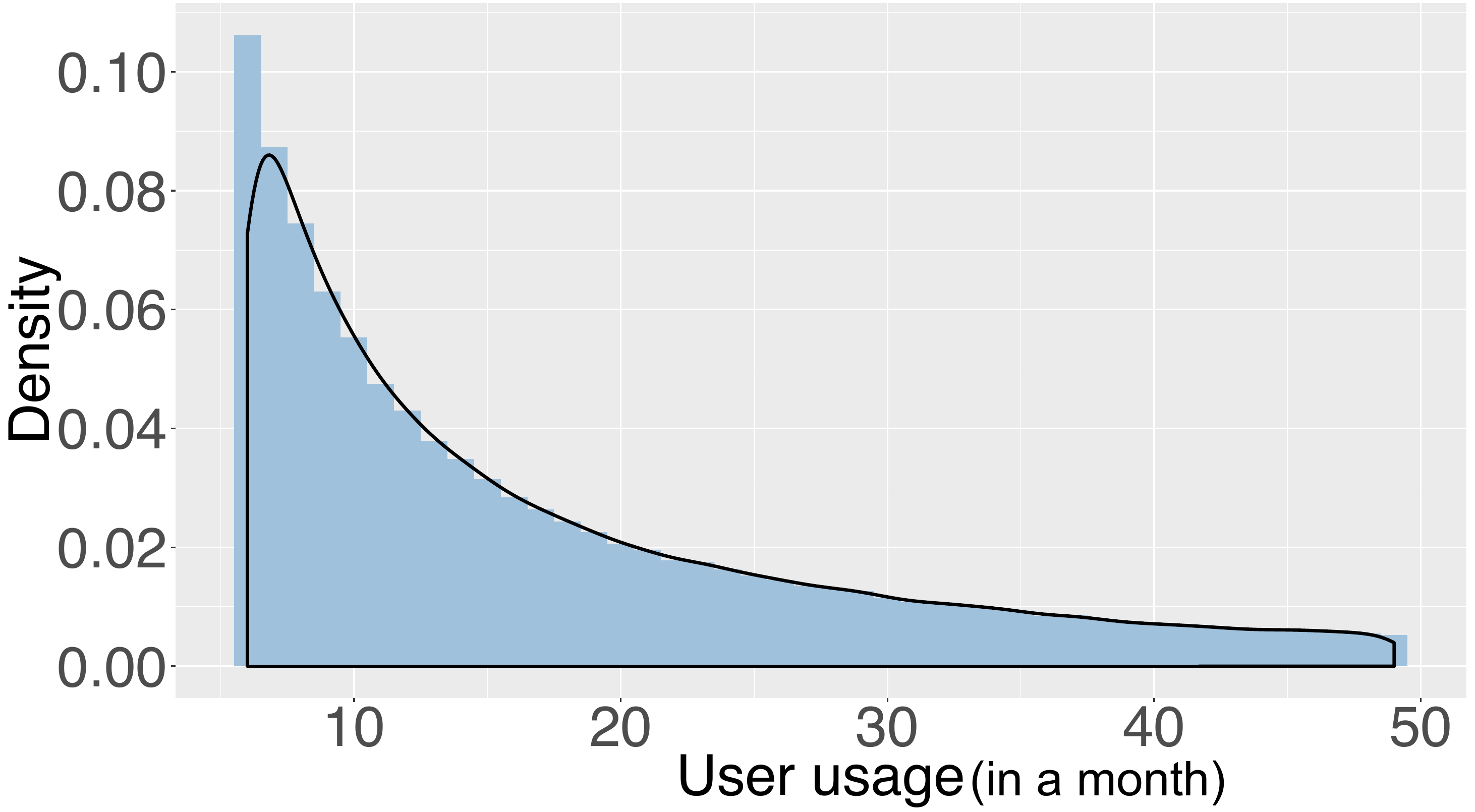}
  \caption{}
 \label{fig:user-usage}
 \end{subfigure} 
 \begin{subfigure}{0.61\columnwidth}
 \includegraphics[width=\textwidth]{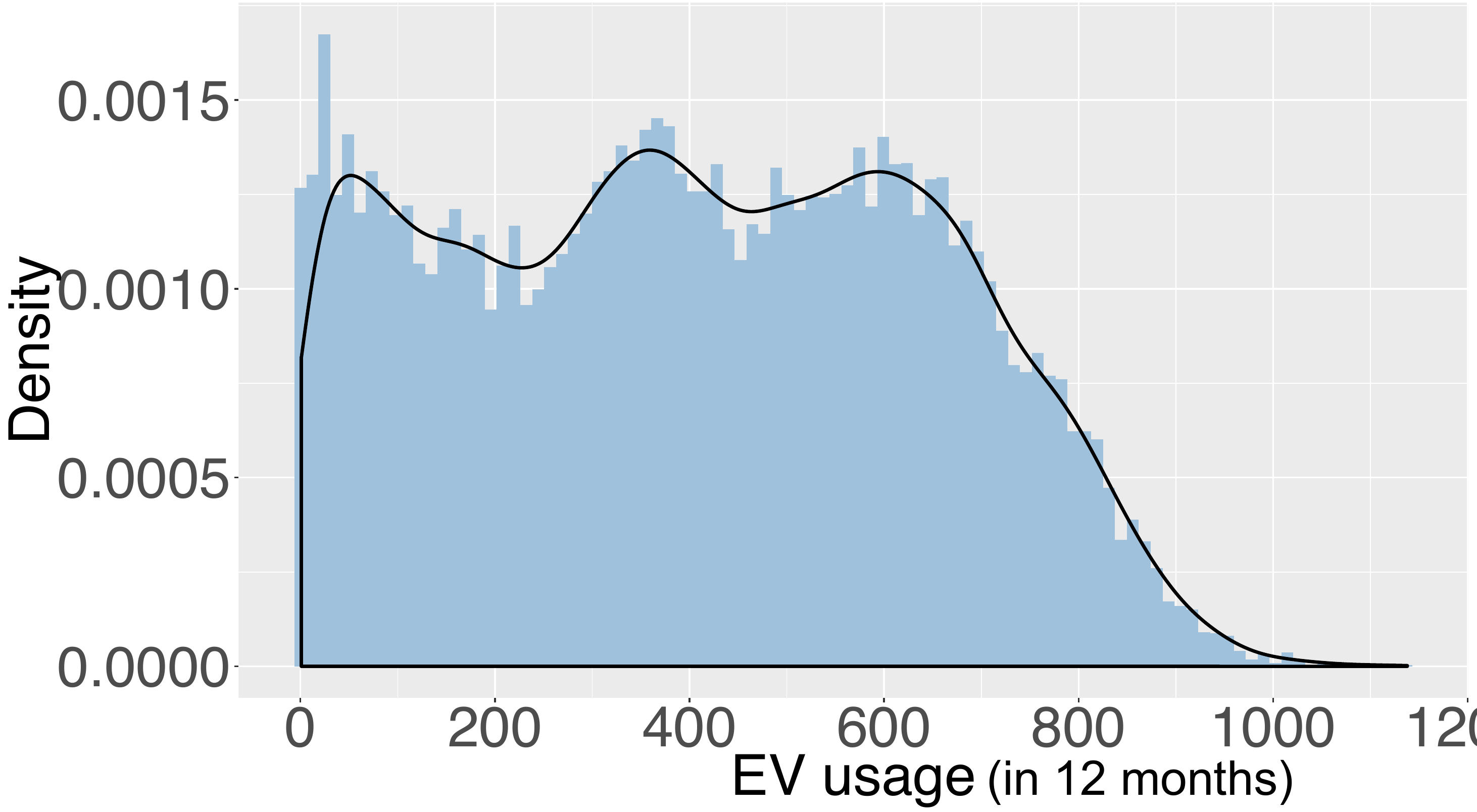}
 \caption{}
 \label{fig:ev-usage}
 \end{subfigure} 
 \caption{The distributions of (a) the order duration, (b) the monthly usage frequency per user, and (c) the overall EV utilization.}
 \label{fig:patterns}
  \vspace{-5mm}
 \end{figure*}

\noindent \textbf{Demand Patterns. }
We observe that the demand patterns of the shared e-mobility system also vary significantly across space and time. Firstly, there are clearly different temporal patterns between weekdays and weekends. As shown in Fig.~\ref{fig:week}, we see the two peaks on weekdays align well with the morning (7-9am) and evening (5-7pm) rush hours in Shanghai. It is also interesting to see that the demand at evening peaks are higher than mornings. One possible reason is that people choose not to use shared EVs to avoid congestion during journeys to work, while are more flexible and willing to drive when finish in evenings. Intuitively during those rush hours the rebalancing task is more challenging, where significant demand would surge across the city. For instance, we may have to reposition more EVs during those times to ensure the balance of the system. In addition, Fig.~\ref{fig:patterns-morning} and Fig.~\ref{fig:patterns-evening} show the spatial distributions of demand at both mornings and evenings. We see high volumes of demand are typically generated at suburban areas in the mornings, while at central areas in the evenings, reflecting the typical commuting needs. Therefore, rebalancing strategies should be able to take such spatio-temporal characteristics into account, and make informed decisions as how to reposition EVs in different context.

\noindent \textbf{Usage Characteristics.}
As shown in Fig.~\ref{fig:order-len}, we see that the typical usage of the shared e-mobilty system is for short trips (mean order length 46min). In light of this, user-incentive based rebalancing strategies are more sensible and beneficial than using dedicated stuff, since users typically would not mind repositioning EVs after short trips in exchange for incentives~\cite{singla:AAAI:2015}. We also find that $>$50\% users tend to use the system for less than 30 times in a month, i.e. roughly on daily basis, as shown in Fig.~\ref{fig:user-usage}. This implies that although with incentives, rebalancing strategies still need to be carefully designed to not require excessive user efforts. Interestingly, we also find the EV utilization across the system is skewed. Fig.~\ref{fig:ev-usage} shows the distribution of usage frequency of EVs. We see that some EVs have been used more than the others, e.g. they might be deployed to popular stations such as airports. This will lead to imbalanced vehicle conditions across the fleet, e.g. they may suffer from early battery degradation or over wear and tear, which ideally should be avoided. Rebalancing can naturally alleviate this, where EVs are often redistributed from popular stations to quieter ones, smoothing out the overall aging of the entire fleet.

% Incentive-based Rebalancing
\subsection{The Fleet Rebalancing Problem}
\label{sub:problem}
In this paper, we consider fleet rebalancing by incentivizing the users. Let $o_t = (s^{\text{o}}, s^{\text{d}})$ be an order placed by a user at time $t$, requesting to rent a vehicle from station $s^{\text{o}}$ and return to $s^{\text{d}}$. As discussed above, due to the imbalance of user demand over space and time, during operation the fleet can become skewed across the system, where some stations are too crowded (i.e., not enough places to park/charge), while the others are depleted (i.e., not enough EVs to rent). In general, the task of rebalancing is to suggest alternative stations for either pick up or return, or both, of an order $o_t$, so that collectively the distribution of fleet is more balanced for future operations. Note that here rebalancing does not seek to normalize fleet distribution over space and time, but rather to improve the long-term system performance, e.g. satisfying more user demand and/or achieving higher revenue. Without loss of generality, in this paper we consider rebalancing operations at destinations, i.e. to reposition the EV serving the current order $o_t$ to an alternative station $s^{\text{d}'}$ instead of the original $s^{\text{d}}$, if the remaining EV range is sufficient. We motivate the users to perform this rebalancing operation by offering monetary reward $v(s^{\text{d}}, s^{\text{d}'})$, depending on the extra distance incurred.
The user may or may not accept the offer according to a pre-defined user model~\cite{singla:AAAI:2015}. In this work we consider an \textit{incentive acceptance probability} $p$, which describes the likelihood of users accepting the incentives for rebalancing. If offer accepted, we pay the reward directly, e.g., discounting the order price, while otherwise we allow the users to return the EV to the original destination and complete the order normally.

Therefore, the rebalancing problem studied in this paper is that given the total available budget $B$ on user incentives, for each order $o_t = (s^{\text{o}}, s^{\text{d}})$, we want to decide where to reposition the EV to minimize the future customer loss (i.e. satisfying as much user demand as possible) while maximizing the net revenue of the shared e-mobility system, given limited EV range, typical EV charging time, and the dynamically expanding station network.

%%%%%%%%%%%%%%%%%%%%%%%%%%%%%%%%%%%%%%%%%%%%%%%%%%%
%% Method
%%%%%%%%%%%%%%%%%%%%%%%%%%%%%%%%%%%%%%%%%%%%%%%%%%%

\section{Rebalancing the Shared e-Mobility Systems}
\label{sec:method}

\subsection{Fleet Rebalancing as a Multi-agent RL Task}
\label{sub:method-formulation}
We model the above fleet rebalancing problem with the multi-agent reinforcement learning (MARL) formulation. In practice, MARL assumes that multiple agents coexist in a shared environment, where each agent operates based on the rewards it receives. Intuitively, individual agents aim to advance their own rewards, which may align or be opposed to the interests of the others, leading to non-trivial cooperative or competitive behaviours~{\cite{marl}}. Concretely, we consider a Markov Game $G = (N, \mathcal{X}, \mathcal{A}, \mathcal{T}, \mathcal{R}, \gamma)$, where at most $N$ agents interact with the environment, characterized by the states $\mathcal{X}$ and transition functions $\mathcal{T}$. $\mathcal{A}$ is the joint actions of the agents, $\mathcal{R}$ is the reward functions, and $\gamma$ is the discount factor. Comparing to single agent settings, this multi-agent formulation is more suitable for our problem, as multiple agents can better exploit the decentralized nature of the rebalancing task, and by design it allows new agents to join in a flexible way, capturing the dynamic system expansion in our context. In the following, we explain the elements of our multi-agent reinforcement learning (MARL) formulation for EV fleet rebalancing in more detail. 

\begin{figure}[t]
\centering
\includegraphics[width=\columnwidth]{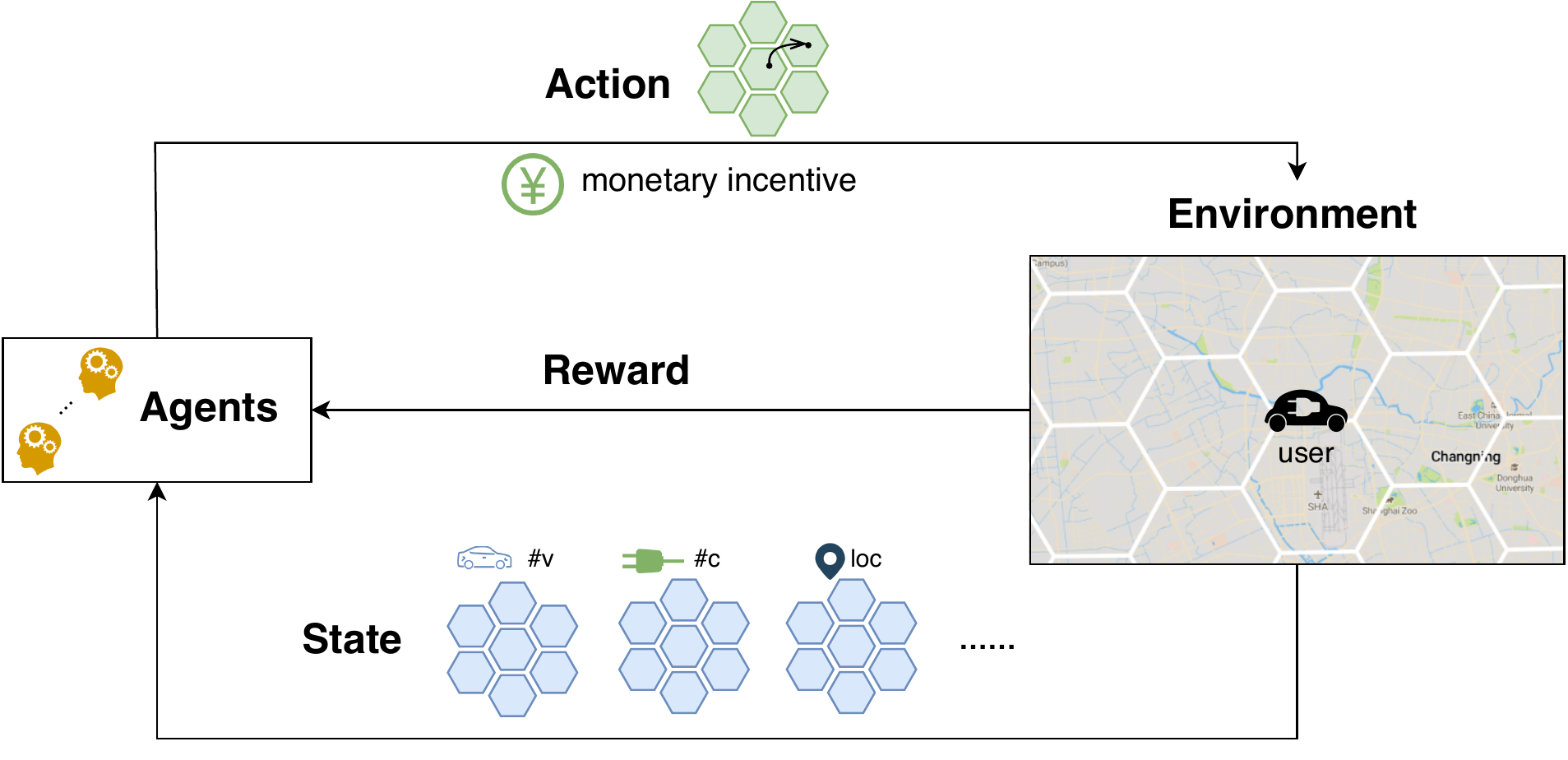} 
\caption{An illustration of the MARL formulation for user-incentive based EV fleet rebalancing.}
\label{fig:marl-flow}
\vspace{-5mm}
\end{figure}

\noindent \textbf{Agents: } 
We assume an agent controls the operation of EV stations within a hexagonal region, as shown in Fig.~{\ref{fig:marl-flow}}. As the shared e-mobility system is continuously expanding, in our case, we assume at time $t$ there are $N_t$ grids managed by the agents, while new grids can be directly appended when needed. In the following text, for simplicity we assume the maximum number of grids (also agents) is $N$, which is set to be large enough to cover the entire city.

\noindent \textbf{States: }
At $t$, the global state $\boldsymbol{x}_t$ is the combination of states for each hexagonal grid $\boldsymbol{x}_t = \{x_t^i\}$, $i\in[1, N]$. For the $i$-th grid, state $x_t^i$ encodes information of the shared e-mobility system within the grid boundary. For each station in grid $i$, $x_t^i$ includes its location \#\texttt{loc}, number of available charging docks \#\texttt{c}, number of EVs parked in the stations \#\texttt{v} and their individual range, as well as the potential future rent/return requests (number of orders) and the average value of potential future orders in the next timestamp. Note that these information is provided by the simulator, which will be discussed in more detail in Sec.~{\ref{sec:simulator}}.

\noindent \textbf{Agent Observations: }
Let agent $i$ manage the $i$-th hexagonal grid. At time $t$, we assume the agent makes a \textit{partial} observation of the global state $\boldsymbol{x}_t$, drawing from the grids within its \textit{two-hop neighbourhood}, i.e. normally it observes the states of itself and the 18 grids around it. This enables the agent to interact with their neighbouring agents, and learn to cooperate. In addition, this setting also avoids unnecessary computation, i.e. an agent doesn't need to track the states of far away peers, whose actions should have little impact on itself. 

\noindent \textbf{Actions. }
For agent $i$, action $a_t^i$ describes its rebalancing strategy at time $t$, i.e., for each EV $a_t^i$ decides if it should be repositioned and where. Without loss of generality, we assume agents only reposition EVs to stations within the one-hop neighbourhood of the destination grid to avoid excessive user effort. As there are stations being deployed or closed dynamically in the shared e-mobility system, the action space $\mathcal{A}_t^i$ is non-stationary, i.e., the candidates of the reposition stations vary over time. Collectively a joint action $\boldsymbol{a}_t \in \mathcal{A}_t^1 \times ... \times \mathcal{A}_t^{N}$ specifies the EV rebalancing strategies of all $N$ agents.

\noindent \textbf{State Transitions. }
The state transition probabilities $\mathcal{T}$ are defined as $\mathcal{T}(\boldsymbol{x}_{t+1}|\boldsymbol{x}_{t}, \boldsymbol{a}_t, \boldsymbol{u}_t)$, where $\boldsymbol{x}_{t}$ is the previous state, $\boldsymbol{a}_{t}$ is the joint action of the agents, and $\boldsymbol{u}_{t}$ describes the dynamics caused by the station network expansion happened at $t$, i.e., which new stations are deployed with how many new EVs, and which exiting stations are off-line from time $t$. Therefore in our case the environment is non-stationary, as the state transitions are induced not only by the actions $\boldsymbol{a}_{t}$ of the agents, but also the external input $\boldsymbol{u}_{t}$, which is often governed by a random process.

\noindent \textbf{Reward Function. }
In our formulation, the collective reward of taking joint action $\boldsymbol{a}_t$ given the current state $\boldsymbol{x}_t$ is determined by a reward function: 
\begin{equation}
    \label{eq:general-r}
    \mathcal{R}(\boldsymbol{x}_t, \boldsymbol{a}_t): \mathcal{X}\times \mathcal{A}_t^1\times ...\times \mathcal{A}_t^{N} \rightarrow \mathbb{R}
\end{equation}
where for each agent, we consider the following reward $r_t^i$:

\begin{equation}
    \label{eq:station-r}
    r_t^i = g_t^{\text{d}'} + \alpha_1 v_t^{\text{d}'} + \alpha_2 b_t^{\text{d}'} - \alpha_3 d(s^{\text{d}'}, s^{\text{d}})
\end{equation}
Here $g_t^{\text{d}'}$ is the expected demand gap at station $s^{\text{d}'}$ in the next timestamp, i.e., the number of orders minus the number of available EVs onsite, and $v_t^{\text{d}'}$ is the expected average order value at station $s^{\text{d}'}$. $b_t^{\text{d}'}$ is a binary variable indicating if station $s^{\text{d}'}$ is empty, i.e., there is no EV in the station at $t$. We use $b_t^{\text{d}'}$ to explicitly encourage agents to position EVs to those empty stations, which are very likely to cause unsatisfied demand in the future. The penalty term $d(s^{\text{d}'}, s^{\text{d}})$ is the cost we pay (monetary reward to the user) for this reposition, which is proportional to the squared distance that one has to travel from the original $s^{\text{d}}$ to the new destination station $s^{\text{d}'}$. 
The weights $\alpha_1$, $\alpha_2$ and $\alpha_3$ scale the different reward/penalty terms to approximately the same range, which are determined empirically via grid search. In our experiments we set $\alpha_1$ = 1, $\alpha_2$ = 2 and $\alpha_3$ = 0.3. Given the reward function, each agent aims to maximize its discounted reward $\mathbb{E} [ \sum_{k=0}^{\infty} \gamma^{k} r_{t+k}^i ]$, where $\gamma \in [0,1]$ is the discount factor. Fig.~{\ref{fig:marl-flow}} shows an illustration of the MARL formulation of the rebalancing task.

As discussed above, due to the fact that the shared e-mobility system is dynamically expanding over time, the action spaces of the agents are non-stationary. This means in both training and testing, typical RL techniques such as Q-learning or policy-based approaches are not directly applicable, as they often require a fixed set of actions to evaluate their Q values or the probability distribution over them. In the following, we first discuss the standard policy optimization approach that is designed for stationary action spaces. We explain how it can discover the optimal policy in this setting, and then we present the proposed approach which extends the standard method with action cascading to handle non-stationarity.

\subsection{Standard Policy Optimization}
\label{sub:method-standard}

Depending on specific learning tasks, there are two main types of model-free MARL algorithms in practice: value-based and policy-based. In this paper, we consider the recent policy optimization approach~{\cite{konda:NIPS:2000}}, which iteratively searches for better policies that yield larger returns.

More concretely, let $\pi_{\theta}$ be the policy parameterized by $\theta$, which is often implemented as policy networks. In our case, the agents aim to maximize the expected discounted reward since the beginning of time: $\eta (\pi_{\theta}) = \mathbb{E} \left[ \sum_{t=0}^{\infty} \gamma^{t}\boldsymbol{r}_t \right]$, where $\boldsymbol{r}_t = \mathcal{R}(\boldsymbol{x}_t, \boldsymbol{a}_t)$ is the joint reward. In practice, $\eta (\pi_{\theta})$ can be optimized by the Minorize-Maximization (MM) algorithm~\cite{hunter:TAS:2004}, which tries to find a surrogate function approximating the lower bound of $\eta$ at the current policy $\pi_{\theta}$ and optimize it iteratively. In particular, we consider the following objective function $L$:
\begin{equation}
    \label{eq:l-func}
    L(\theta) = \hat{\mathbb{E}} \left[ \frac{\pi_{\theta} (\boldsymbol{a}_t | \boldsymbol{x}_t)}{\pi_{\theta_{\text{old}}} (\boldsymbol{a}_t | \boldsymbol{x}_t)} \hat{A}_{t}\right]
\end{equation}
where $\boldsymbol{a}_t$ and $\boldsymbol{x}_t$ are the joint actions and states at time $t$, and $\hat{\mathbb{E}}[\cdot]$ is the empirical average over the batch of samples. $\hat{A}_{t}$ is an estimator of the advantage function $A_t$, which is the benefit of taking a specific action $\boldsymbol{a}_t$ under the current state $\boldsymbol{x}_t$ than the expected state value: 
\begin{equation}
    \label{eq:adv-func}
    A_t (\boldsymbol{a}_t, \boldsymbol{x}_t) = Q (\boldsymbol{a}_t, \boldsymbol{x}_t) - V (\boldsymbol{x}_t)
\end{equation}
$Q (\boldsymbol{a}_t, \boldsymbol{x}_t)$ is the $Q$-function and $V (\boldsymbol{x}_t)$ is the value function~{\cite{konda:NIPS:2000}}. The intuition is that $L$ approximates $\eta (\pi_{\theta})$ locally at the current policy $\pi_{\theta}$, but can get inaccurate as it moves away from $\pi_{\theta}$. Therefore, to avoid updating the policy too much, constrained forms of $L(\theta)$ are often considered, such as using KL divergence or clip functions~\cite{schulman2017proximal}. We then iteratively optimize the objective function, until the optimal policy $\pi_{\theta^{*}}$ can be found. A key assumption of standard policy optimization is that the action spaces of the agents are stationary, i.e., at each timestamp the learned policy $\pi_{\theta}$ outputs a distribution over the fixed set of possible actions $\mathcal{A}$. However this is not directly applicable in our case, as the candidate stations to which an agent can reposition EVs are fixed but dynamically evolving.

\subsection{Policy Optimization with Action Cascading}
\label{sub:method-proposed}
We now present the proposed policy optimization approach with action cascading (ac-PPO), which extends the standard algorithms and is able to handle the non-stationarity in our EV fleet rebalancing problem. The key intuition is that in our settings the action of repositioning an EV to an alternative station can be viewed as a sequence of two sub-actions, where we first decide which grid the EV should go to, and then figure out which station within that selected grid should be the target. In essence, we chain two sub-actions, one inter-grid and the other intra-grid, to achieve the desired goal of repositioning the vehicle. One of the benefits of this action cascading is that now the inter-grid actions can have fixed action spaces, while the non-stationarity of the station network would only affect the intra-grid actions. This makes it possible to fit our problem into the policy optimization framework discussed above, where the non-stationarity within different grids can be handled by separate policy selectors. In the following, we first explain the design of action cascading in more detail, and then we show how we adapt the reward structure to stabilize training. 

\begin{figure}[t]
\centering
\includegraphics[width=\columnwidth]{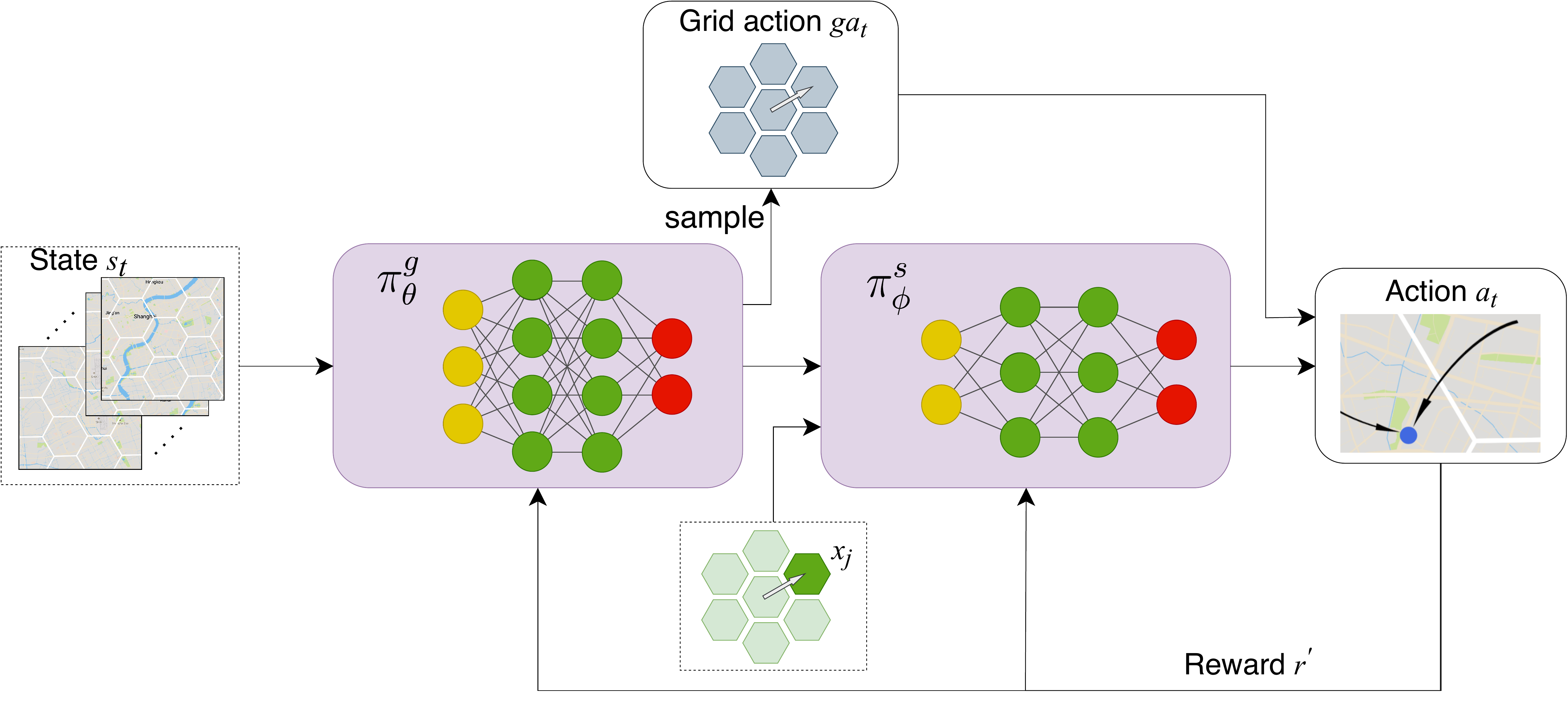} 
\caption{Overview of the proposed action cascading approach.}
\label{fig:policynet}
\vspace{-5mm}
\end{figure}

\noindent \textbf{Action Cascading. }
Let $a_t^i$ be the action of agent $i$ at time $t$. We assume $a_t^i$ can be decomposed as $a_t^i = (ga_t^i, sa_t^i)$, where $ga_t^i$ is the inter-grid action that decides which grid within the neighbourhood the EV should be redirected to, and $sa_t^i$ is the intra-grid action which determines the actual destination station within the selected grid. Clearly, here $ga_t^i$ has a fixed action space, which contains the six neighbors around the grid $i$ and itself. Therefore, $ga_t^i$ can be sampled from the output of a standard policy network $\pi_{\theta}^g$ as discussed in Sec.~\ref{sub:method-standard}. Let us assume that we have a $ga_t^i$ that would redirect the EV to a nearby grid $j$. 

Now we need to find the intra-grid action $sa_t^i$ that selects a suitable station within the grid $j$. In this case the action space of $sa_t^i$ is not stationary, since there are always stations deployed or closed in grid $j$. We address this by using an action-in policy network $\pi_\phi^s$, as shown in Fig.~{\ref{fig:policynet}}. Concretely, $\pi_\phi^s$ takes the state $x_t^j$ and the agent observation $o_t^j$ of grid $j$, together with the output of the last layer of the inter-grid policy network $\pi_{\theta}^g$ as input. Here the state $x_t^j$ and observation $o_t^j$ encodes information about the currently available stations and vehicles within grid $j$, and the output from $\pi_{\theta}^g$ conditions $x_t^j$, i.e. providing context from the precedent inter-grid action. The output of the network $\pi_\phi^s$ are deterministic values of each station within grid $j$, indicating the ``fitness'' of those stations if chosen as the destination of repositioning, and we select the one with highest value as action $sa_t^i$. Therefore, $\pi_\phi^s$ ranks the currently available stations in grid $j$, based on the information from $\pi_{\theta}^g$ and the current states/observations, and determines $sa_t^i$ accordingly. 

Essentially, we use two policy networks that are connected, to compute the inter-grid and intra-grid actions respectively. During training, we only sample from the inter-grid policy network $\pi_{\theta}^g$, while considering the intra-grid policies of $\pi_\phi^s$ are deterministic, which makes the training more data efficient. In our implementation, we train the networks with the following clipped objective function: 

\begin{equation}
\label{eq:clip-l}
L^\text{CLIP}(\theta,\phi)  = \hat{\mathbb{E}} \left[ \text{min}( R_t^\theta \hat{A}_{t}^{\theta, \phi},\text{Clip} (R_t^\theta, 1-\epsilon, 1+\epsilon)\hat{A}_{t}^{\theta, \phi} \right]
\end{equation}
where $R_t^\theta$ is the probability ratio between the new and old inter-grid policy: 
\begin{equation}
\label{eq:clip-r}
R_t^\theta = \frac{\pi_{\theta}^g (\boldsymbol{ga}_t | \boldsymbol{x}_t)}{\pi_{\theta_{\text{old}}}^g (\boldsymbol{ga}_t | \boldsymbol{x}_t)}
\end{equation}
This $R_t^\theta$ together with the Clip function constrains the policy updates to avoid obtaining very different new policies. The $\epsilon$ in Eq.~{\eqref{eq:clip-l}} is a hyperparameter, which is usually set to values around 0.2$\sim$0.3, controlling the threshold of the Clip function. Note here we only consider the inter-grid policy $\pi_{\theta}^g$, since the output $\boldsymbol{ga}_t$ has stationary action space. On the other hand, the advantage function $\hat{A}_{t}^{\theta, \phi} = Q (\boldsymbol{a}_t, \boldsymbol{x}_t) - V(\boldsymbol{x}_t)$ considers both inter and intra-grid policies, since the reward $\boldsymbol{r}_t$ is given to the full actions $\boldsymbol{a}_t = (\boldsymbol{ga}_t, \boldsymbol{sa}_t)$, where the $Q$ function is evaluated with the discounted rewards of the actions obtained in this experience.

\noindent \textbf{Reward Regularization. }
Essentially, the proposed ac-PPO addresses the non-stationarity in action spaces by decomposing the action into the sequence of inter-grid and intra-grid sub-actions, and using two connected policy networks to determine them. Therefore, we fit the non-stationary rebalancing problem into the policy optimization framework, by allowing non-stationary reward functions. In fact, from the view of the inter-grid policy network $\pi_{\theta}^g$, the reward distribution of the same action (e.g., repositioning the EV to the grid directly above) across different timestamps may be different, because the set of stations within the grid are changing, and the intra-grid policy network $\pi_{\phi}^s$ is very likely to select different destination stations. When the station network is very dynamic, such non-stationarity in reward could lead to large gradient variance when training $\pi_{\theta}^g$. To address that, we propose to regularize the reward function $r_t^i$ in Eq.~\eqref{eq:station-r} with a baseline: 
\begin{equation}
    \label{eq:reg-r}
    r_t^{i'} = r_t^i + \beta \bar{r}_t(j)
\end{equation}
where the regularization term $\bar{r}_t(j) = \bar{v}_t(j) \cdot \bar{g}_t(j)$ is the product of the mean order value $\bar{v}_t(j)$ and the average future demand gap $\bar{g}_t(j)$ (\# of user demand - \# of available EVs ) per station in grid $j$, assuming that the action is to reposition the EV to a station in the target grid $j$. Intuitively, $\bar{r}_t(j)$ can be viewed as the ``potential'' of the grid, indicating how much extra revenue one would expect to get if more EVs are repositioned to stations within the grid. In practice, $\bar{r}_t(j)$ is updated every timestamp and is more stable than $r_t^i$, which depends on the particular destination station selected by the intra-grid policy network $\pi_{\phi}^s$. The weight $\beta$ scales the regularization term to adjust its impact during learning, and in our experiments we empirically set $\beta$ = 0.8.

%%%%%%%%%%%%%%%%%%%%%%%%%%%%%%%%%%%%%%%%%%%%%%%%%%%
%% Simulator
%%%%%%%%%%%%%%%%%%%%%%%%%%%%%%%%%%%%%%%%%%%%%%%%%%%

\section{Simulation Design}
\label{sec:simulator}
To support training and evaluation of the proposed MARL algorithm, we design a simulator which captures the dynamics of the environment as well as the interactions between the agents. The simulator is calibrated with one-year real-world data collected from a shared e-mobility system as discussed in Sec.~{\ref{sec:ev-sharing}}.

\begin{figure}[t]
\centering
\begin{subfigure}{0.5\columnwidth}
\includegraphics[width=\textwidth]{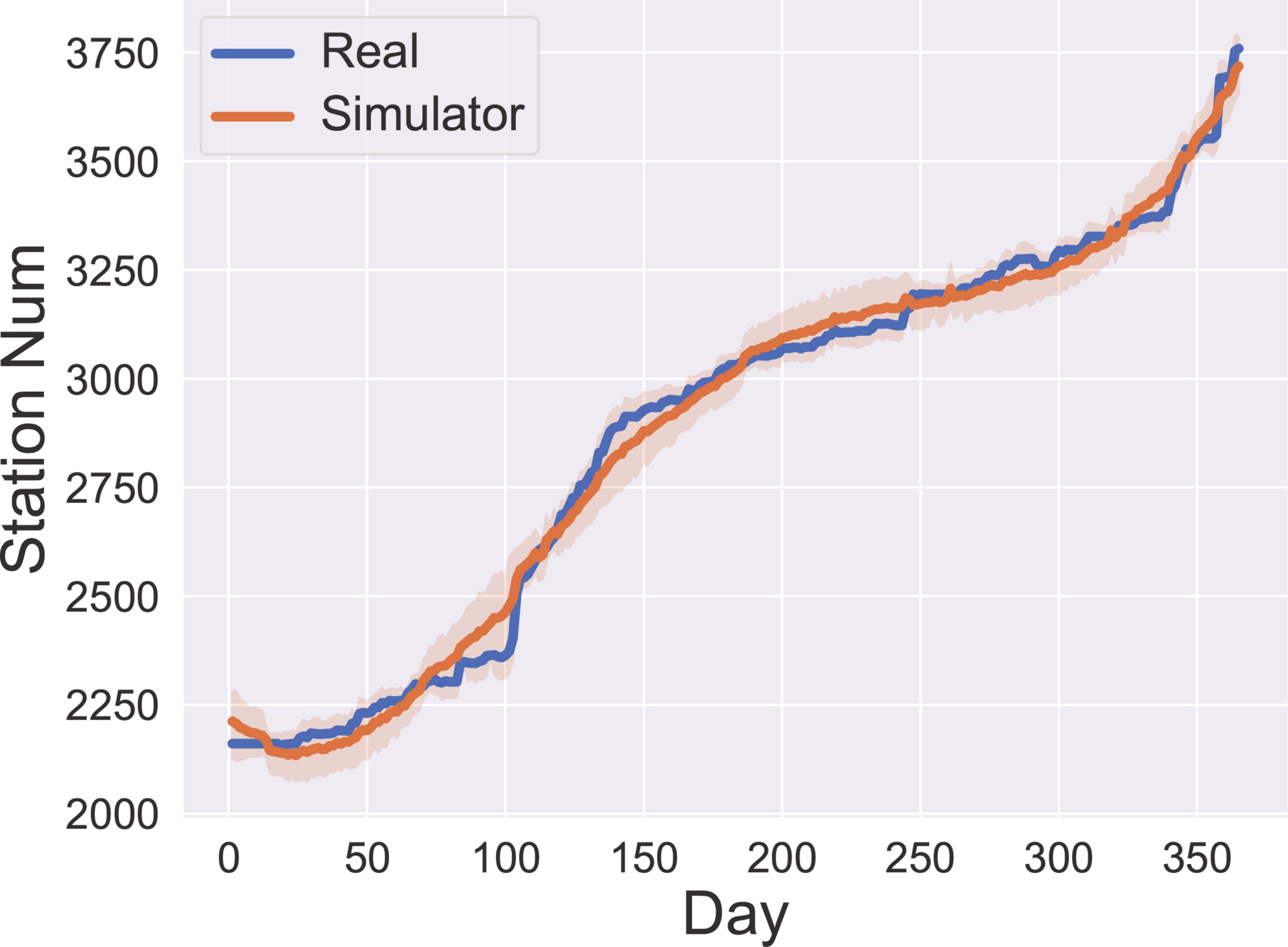}
% \caption{Station Network Expansion}
\caption{}
\label{fig:calibration-station}
\end{subfigure}~\hspace{0mm}
\begin{subfigure}{0.5\columnwidth}
\includegraphics[width=\textwidth]{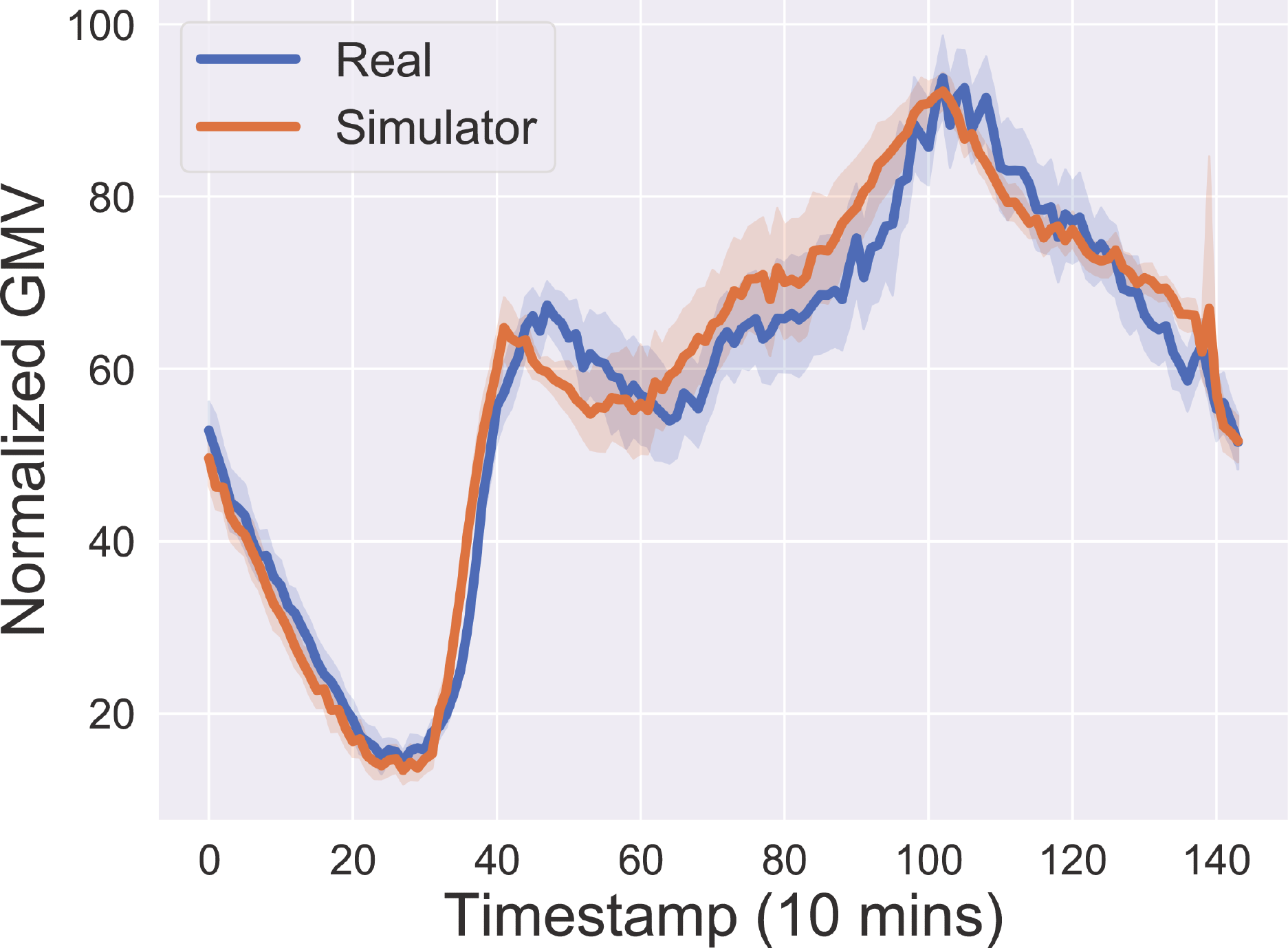}
% \caption{Gross Merchandise Value}
\caption{}
\label{fig:calibration-gmv}
\end{subfigure}
\caption{Simulator calibration and fidelity. (a) Simulated vs. real station network expansion for one year (averaged over 10 runs), and (b) Simulated vs. real Gross Merchandise Value (GMV) for one day (averaged over 7 days).  } 
\vspace{-5mm}
\end{figure}

\noindent \textbf{Basic Settings. }
In our simulator, we set 10 mins as one timestamp, i.e., a day (24 hours) contains 144 timestamps. The space is partitioned into hexagonal grids, where each agent manages one grid and can reposition EVs to the neighbouring 6 grids. We follow a similar setup as in previous works (e.g.~{\cite{Li:WWW:2019}}), and set the grid size to 6\textit{km}. In total we have 598 grids covering the entire Shanghai, in which 280 grids are valid since the rest covers areas without road access such as parks or water. To simulate the dynamic system expansion, our simulator uses a random process to control i) the expansion speed, i.e., numbers of new stations to be deployed and existing stations to be closed at any timestamp; and ii) the expansion plan, i.e. where to deploy the new stations, and which existing stations should be closed. The parameters of the random process are learned from the collected expansion data (as in Sec.~{\ref{sec:ev-sharing}}) via the calibration process discussed later. For a new station to be deployed, the simulator sets the number of charging docks \#\texttt{c} and available EVs \#\texttt{v} based historical average from the expansion data. 

\noindent \textbf{User Demand.} 
To simulate the user demand, we train a neural network which takes the current time and the station network as input, and generates the demand distribution over all currently online stations. Concretely, we consider a graph-based approach as in our previous work{~\cite{luo2020d3p}}. To generate a potential order which is a pair of pick up and return stations, we first sample the demand distribution to determine the origin station $s^o$. For a given origin $s^o$, we look at the distribution of destinations in all historical orders which were originated from $s^o$ during similar time periods, and sample from that distribution to determine the destination station $s^d$. Note that the generated user order will only be accepted if at the origin station $s^o$ there is at least one EV with enough range to cover the trip to $s^d$, where the trip time is also sampled from the distribution learned from the historical order data. At initialization, the simulator assumes all the EVs in the system are fully charged. As the simulation progresses, the simulator keeps track of the range of each EV by applying a discharging model~\cite{Tremblay:Charging:2009} when serving orders. Once the order is completed, i.e., the EV is returned to a station and charging, the remaining range of the EV is estimated using a charging model~\cite{Tremblay:Charging:2009}. Of course one could simply add random noise to historical orders to simulate the user demand, but this approach has many limitations. In fact, the added noises are merely linear perturbations to the existing data, while a simulator calibrated in such a way may easily lead to over-fitting when training deep reinforcement learning algorithms. More importantly, this naive method implicitly assumes that the pick-up and return demand is independent, which is often not true in practice. Our approach, on the other hand, can generate synthetic pick-up and return demand pairs that are not in the historical data at all, but exhibit similar statistical properties with the real-world data, as shown above. In addition, it is more flexible to be extended to incorporate additional data modalities when generating user demand, such as weather, social events etc., which could further improve simulation fidelity.

\noindent \textbf{Rebalancing Operations.} 
When there is a need to reposition an EV, our simulator computes an offer of monetary reward, by discounting the original price of that order. The amount of the discount depends on the square of the extra distance that the user has to travel{~\cite{singla:AAAI:2015}}, and is limited by an upper bound set by the shared e-mobility system. Here we consider the point-wise distance, which in our context is a good approximation of the actual travel distance on road networks, yet much more efficient to compute. The simulator assumes that the user would accept this offer according to the incentive acceptance probability $p$, which indicates how cooperative the user is. 
In our experiments we vary $p$ and study how different $p$ values could impact the rebalancing performance. In each of those settings, we sample the user behaviour, i.e. to accept or reject the reposition offer, according to the selected $p$ value. If the offer is accepted, the simulator updates the order information (e.g., discounting the price, changing the destination station), and also updates the status of the EVs and stations accordingly.

\noindent \textbf{Simulator Calibration and Fidelity.}
We calibrate our simulator with the real EV sharing data to ensure its high fidelity. As shown in Fig.~{\ref{fig:calibration-station}}, the patterns of simulated system expansion are very close to the actual expansion exhibited in the real system, with Pearson correlation 0.9957 and $p$ value $p<$1e-10. For demand generation, we use the calibrated system expansion dynamics, and further tune the simulator with respect to the Gross Merchandise Value (GMV), indicating the total revenue of the system. Fig.~{\ref{fig:calibration-gmv}} shows that the simulated order data has very similar behaviour in GMV with the real data, where the Pearson correlation between simulated and real GMV is 0.9599 with $p$ value $p<$1e-10.

%%%%%%%%%%%%%%%%%%%%%%%%%%%%%%%%%%%%%%%%%%%%%%%%%%%
%% Eval
%%%%%%%%%%%%%%%%%%%%%%%%%%%%%%%%%%%%%%%%%%%%%%%%%%%
%!TEX root = main.tex

\section{Evaluation}
\label{sec:exp}

\subsection{Experimental Settings and Baselines}
\label{sub:exp-setting}
We adopt the simulator developed in Sec.~{\ref{sec:simulator}} for both training and testing, which is calibrated with the data collected from a real-world shared e-mobility system as discussed in Sec.~{\ref{sub:ev-data}}. In particular, we use the station deployment data over a year to simulate the system expansion process, and the data of historical orders to train a neural network to synthesize the user demand over space and time (details explained above in Sec.~{\ref{sec:simulator}}). 
We compare the proposed approach with the following baselines:
\begin{itemize}[leftmargin=5mm, topsep=0pt]
     \item \textbf{No Rebalancing (NR)}, which simulates the operation of shared e-mobility system without any rebalancing actions.
    \item \textbf{Random Rebalancing (RND)}, where in rebalancing the EVs are repositioned randomly to nearby stations (with at least one charging dock available) within a certain radius of the original destinations.
    \item \textbf{Revenue Greedy (REV)}, which is similar with RND but selects the stations with the highest average order values.
    \item \textbf{Demand Gap Greedy (DMD)}, which prefers the stations with the highest demand gap in the vicinity.
    \item \textbf{STRL}, which is our implementation of the the state-of-the-art rebalancing approach~\cite{li:Kdd:2018} for shared mobility systems. It uses multi-agent spatial-temporal reinforcement learning to reposition shared bikes across different stations. 
\end{itemize}

To further validate the performance of the proposed action cascading, we also consider different variants of our RL algorithm for ablation study. For the inter-grid actions (i.e., determining which grid for repositioning), we consider the following different implementations of the policy network $\pi_{\theta}^g$ (as explained in Sec.~\ref{sub:method-proposed}): 
\begin{itemize} [leftmargin=5mm, topsep=0pt]
    \item \textbf{Policy Gradient (ac-PG)}, which uses the standard policy gradient technique to determine the inter-grid actions. The policy network is implemented with a four layer MLP, and we use learning rate 4e-4.
    \item \textbf{Deep Q Networks (ac-DQN)}, which uses a $Q$-network to approximate the action-state values. In our implementation, we use four layer MLP and $\epsilon$-greedy policy as the agent policies, where $\epsilon$ is annealed from 0.1 to 0.02 in training. The learning rate is set to 5e-4.
    \item \textbf{Advantage Actor Critic (ac-A2C)}, which uses two separate networks (the actor and the critic) to produce actions and estimate the advantage values respectively. In our implementation, we use two four layer MLPs for the actor and critic network. We use learning rate 1e-4 for both networks.
    \item \textbf{Proximal Policy Optimization (ac-PPO)}, which is the policy optimization approach as discussed in Sec.~\ref{sub:method-proposed}. In particular, we use a four layer MLP as the policy network to generate the inter-grid actions, which also estimates the state values. The learning rate is set to 5e-5.
\end{itemize}

Note that for all the above variants, we use the proposed intra-grid policy network $\pi_{\phi}^s$ as described in Sec.\ref{sub:method-proposed} to determine the later intra-grid actions that redirect the EV to destination stations, and the same function in Eq.~\ref{eq:reg-r} to collect rewards. On the other hand, to evaluate the performance of different approaches for the intra-grid policy network $\pi_{\phi}^s$, we fix the inter-grid policy network $\pi_{\theta}^g$ as the PPO, and consider the following additional variants: 
\begin{itemize} [leftmargin=5mm, topsep=0pt]
     \item \textbf{PPO + Random (PPO+RND)}, which uses PPO to determine which grid to reposition the EV, and then randomly selects a destination station within that grid.
     \item \textbf{PPO + Revenue Greedy (PPO+REV)}, which is similar to the above, but instead of random selection, it finds the destination station with the highest average order values. 
     \item \textbf{PPO + Demand Gap Greedy (PPO+DMD)}, which decides the destination station by finding the one with the largest demand gap within the grid.
\end{itemize}

In our experiments, all the competing approaches are implemented with TensorFlow 1.14.0, and trained with a single NVIDIA 2080Ti GPU. For efficient training, in our implementation all the agents share the same policy and value networks, which also encourage them to collaborate with the others. In practice, the agents can maintain their own network parameters locally, and get updates from a central server. In our ablation study for inter-grid actions, the learning rates of different variants of our approach are set empirically, following the recommended settings as in each deep RL approach and open-source implementations in RLib~{\cite{liang2018rllib}}.

To be fair, we assume that for all approaches the amount of user incentives we have to pay for a particular reposition action is calculated by the same cost model~\cite{singla:AAAI:2015}, which depends on the squared distance between the original destination and proposed reposition station. We evaluate the competing approaches against two main metrics: i) the \textbf{Demand Satisfied Rate (DS)}, which is the percentage of the demand satisfied by an algorithm with respect to the total user demand generated; and ii) the \textbf{Net Revenue Value (NV)}, which is calculated as the GMV of the system subtracts the cost on user incentives.

\subsection{Results}
\label{sub:exp-results}

\begin{table*}[t]
\centering
\small
\begin{tabular}{|@{}c@{}|@{}c@{}||@{}c@{}|c|c|c||c|c|c|c||c|c|c|}
\hline
\multicolumn{1}{|@{}c@{}|}{} & \multicolumn{1}{@{}c@{}|}{NR} & \multicolumn{1}{@{}c@{}|}{RND} & \multicolumn{1}{@{}c@{}|}{REV} & \multicolumn{1}{@{}c@{}|}{DMD}   & \multicolumn{1}{@{}c@{}|}{STRL}  & \multicolumn{1}{@{}c@{}|}{ac-DQN} & \multicolumn{1}{@{}c@{}|}{ac-PG}    & \multicolumn{1}{@{}c@{}|}{ac-A2C}  & \multicolumn{1}{@{}c@{}|}{ac-PPO}  & \multicolumn{1}{@{}c@{}|}{PPO+RND}   & \multicolumn{1}{@{}c@{}|}{PPO+REV} & \multicolumn{1}{@{}c@{}|}{PPO+DMD} \\ \hline\hline
DS   & 74.69\% & 49.79\% & 82.15\% & 81.09\%   & 82.47\% & 83.50\% & 83.64\%  & 85.23\%   & \textbf{88.79\%}  & 53.94\% & 83.19\% & 82.88\% \\ \hline
$\Delta$DS   & --- & -24.90\% & 7.46\% & 6.41\%   & 7.78\% & 8.81\% & 8.95\%  & 10.55\%   & \textbf{14.10\%}  & -20.75\% & 8.51\% & 8.20\% \\ \hline
$\Delta$GMV          & ---   &-36.30\%      &8.25\%  &3.22\%     &9.27\% &10.76\% &11.26\%  &13.13\%    &\textbf{18.13\%}&-29.82\%&10.41\%      &6.22\%                        \\ \hline
$\Delta$NV         & --- & -47.71\% & -7.64\% & -0.48\%   & 1.12\% & 6.95\% & 7.37\%  &8.53\%   & \textbf{12.23\%}   & -48.40\% & -3.27\% &1.72\%        \\ \hline
$\Delta |\boldsymbol{o}|$\slash $|\boldsymbol{a}|$           & --- & --- & 9.28 & 4.98      & 2.08 & 1.14 & 1.09  & 1.11   & 1.12     & --- & 7.82 & 3.53       \\ \hline
\end{tabular}
\caption{Performance of the competing approaches in 1) demand satisfied rate (DS), 2) increased demand satisfied rate ($\Delta$DS) w.r.t. baseline NR, 3) increased \% of GMV ($\Delta$GMV) w.r.t. baseline NR, 4) increased \% of net revenue value ($\Delta$NV) w.r.t. baseline NR, and 5) \# of reposition operations needed to satisfy one extra order ($\Delta |\boldsymbol{o}|$\slash $|\boldsymbol{a}|$, only showing positive values).}
\label{tbl:overall}
\vspace{-5mm}
\end{table*}

\noindent \textbf{Overall Rebalancing Performance. }
The first set of experiments evaluate the overall rebalancing performance of different approaches. Table.~\ref{tbl:overall} shows the demand satisfied rates and the increased net revenue (in percentage) of the competing algorithms. We allow the station network to expand at the normal speed (similar with the real data), where at each timestamp there are new stations deployed and existing ones closed. We can see that comparing to no rebalancing (NR) which is the normal operation, randomly selecting the reposition stations (RND) won't help in neither satisfying user demand, nor improving net revenue: we observe a significant drop in both performance metric. On the other hand, if we are greedy on order values (REV) we do satisfy more user demand by roughly 7\%, but the net revenue actually drop by 8\%. This is because with this algorithm, the agents tend to excessively reposition EVs to those station with high expected order values, while ignore the cost on user incentives. In our experiments, we find that on average, REV would satisfy one extra order (which tends to be of higher values) at the cost of repositioning 9.3 EVs. On the other hand, the demand gap greedy algorithm (DMD) achieves more balanced performance, improving the demand satisfied rate (DS) by 6.41\%, while maintaining similar net revenue with the baseline NR. This is also expected, as sending EVs to stations with larger demand gap is more likely to fulfill future user orders. On average, this DMD has to reposition 4.98 EVs to satisfy one extra order, which is better than REV. We observe that the state-of-the-art STRL outperforms the baselines, with 8\% improvement in DS and 1\% improvement in net revenue (NV). It also has more efficient repositioning as well: on average it repositions 2.1 EVs to satisfy one extra order. It confirms that by using spatial-temporal RL, the STRL can better learn the demand pattern across space and time, and thus make more informed decisions in rebalancing. However, we see that the approaches with the proposed action cascading significantly outperforms STRL. For instance, the best ac-PPO can achieve almost 15\% improvement in demand satisfied rate, while obtaining approximately 12\% more net revenue. This means comparing to the state-of-the-art STRL, our approach can offer two-fold improvement in satisfying user demand, while over 10$\times$ net revenue improvement. In addition, as shown in Table.~{\ref{tbl:overall}}, our approaches, regardless of which RL framework used, have much lower $\Delta |\boldsymbol{o}|$\slash $|\boldsymbol{a}|$ values than the competing approaches. On average, our approaches only need to reposition approximately 1.1 EVs to satisfy an extra order. This means that to achieve similar performance, our approaches require much fewer reposition operations and thus more efficient in general. This validates the effectiveness of the proposed action cascading, and also shows that our approach can cope with the dynamically expanding station network. We will show later that the gap between our approaches and the STRL would be even larger when the system expansion dynamics increases. 

\noindent \textbf{Performance of Inter-grid Policy. }
This experiment compares the performance of different algorithms to learn the inter-grid policy $\pi_{\theta}^g$ in our action cascading framework, which decides the grid that the EVs should be repositioned to. We consider four different learning algorithms, the policy gradient (PG), DQN, A2C and PPO. Note that here we plug in those algorithms to our action cascading framework, while using the same intra-grid policy network $\pi_{\phi}^s$ later and feed the algorithms with the same reward. Firstly, we see that even the weakest performed algorithm ac-DQN can achieve better performance than the state-of-the-art STRL, with approximately 1\% improvement in demand satisfied rate and 5\% increase in net revenue. This is because STRL doesn't have the mechanism of handling station network expansion, while the proposed action cascading can effectively work with the non-stationarity in action space. The reason why ac-DQN performs not so well is because that DQN typically works well when the action space is finite with clear reward structures, while in our case although action cascading can address varying action spaces, the non-stationarity in rewards can still affect the performance of ac-DQN. On the other hand, policy gradient (PG) only performs slightly better than DQN, but is inferior to A2C, which is about 2\% better in both satisfied demand rate and increase in net revenue. This is because in practice the variance of the gradients computed by PG can be large, and thus it is very likely to deviate from the optimal direction if the learning rate is too high. Comparing to ac-A2C, the ac-PPO (discussed in Sec.~\ref{sub:method-proposed}) provides a further improvement of roughly 4\% in both demand satisfied rate and net revenue, achieving $>$14\% better demand satisfied rate and $>$12\% extra net revenue than no rebalancing (NR). This maps to more than 200,000 USD extra revenue per month according to the real data where the mean order value is around 3.8 USD and average number of orders per month is about 500,000. 

\begin{figure}[t]
\centering
\begin{subfigure}{0.5\columnwidth}
\includegraphics[width=\textwidth]{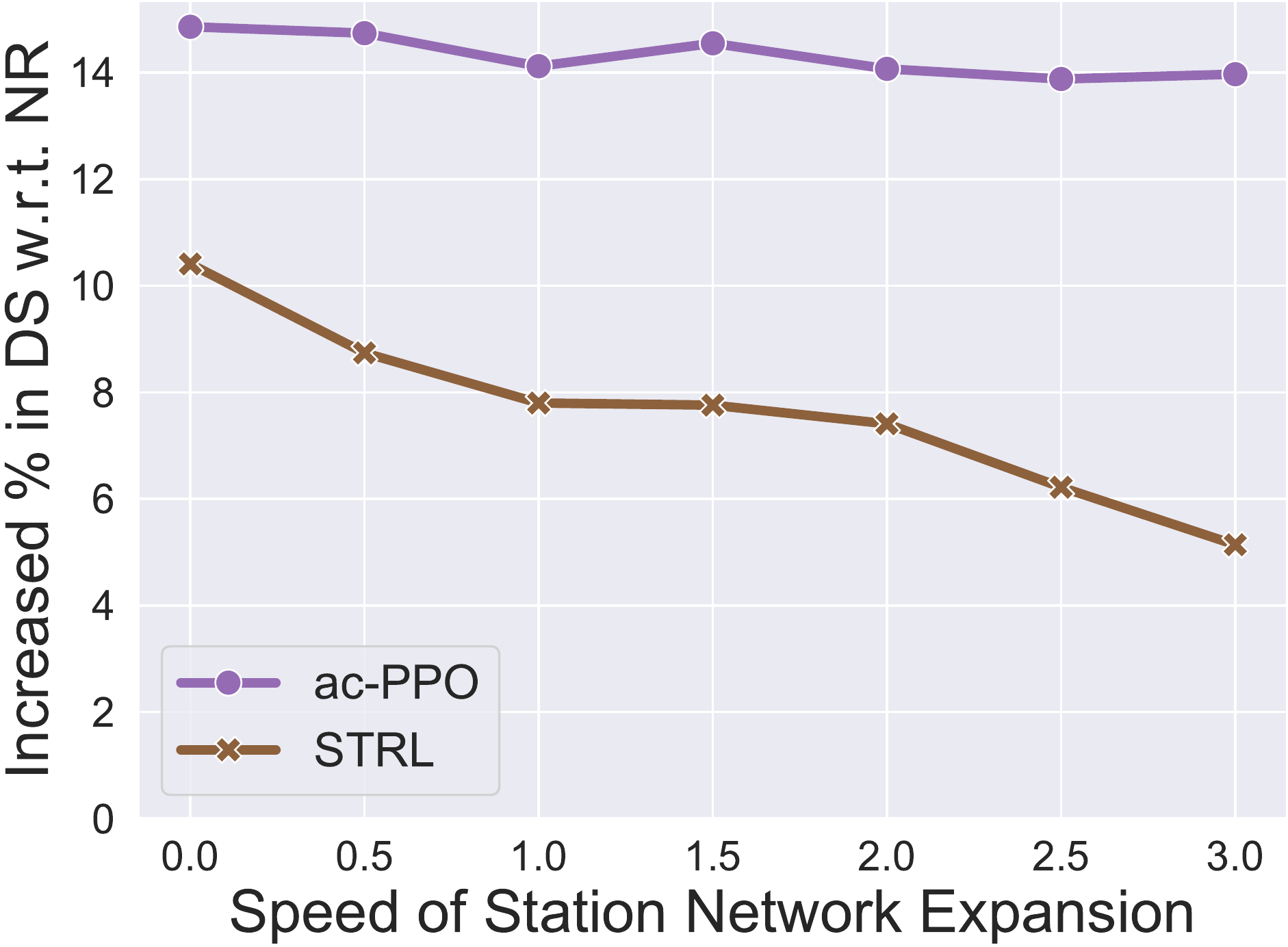}
\caption{Increased \% in DS}
\end{subfigure}%~\hspace{5mm}
\begin{subfigure}{0.5\columnwidth}
\includegraphics[width=\textwidth]{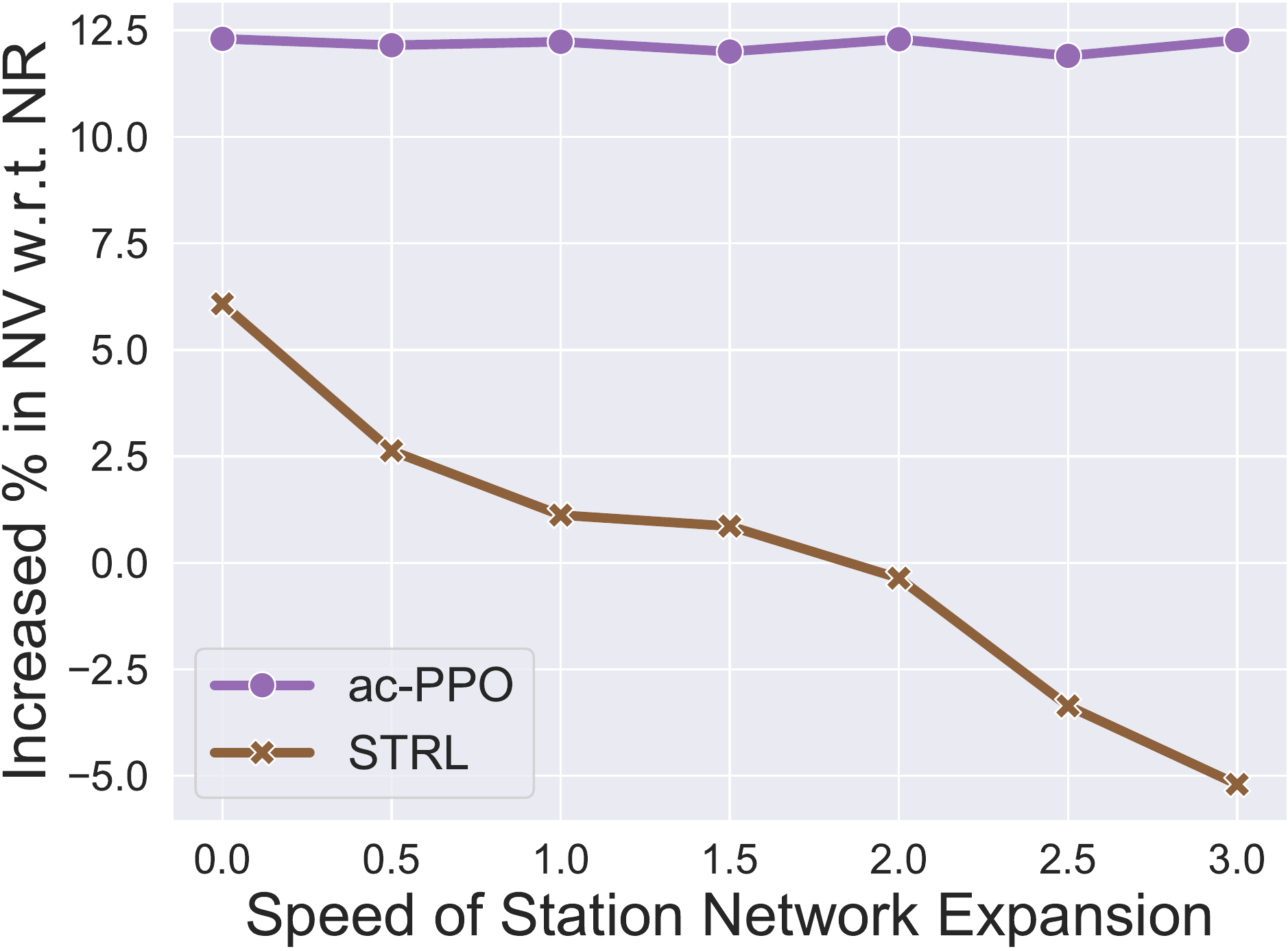}
\caption{Increased \% in NV}
\end{subfigure}
\caption{Performance of the proposed ac-PPO and STRL under different speeds of station network expansion.} 
\label{fig:expansion}
\vspace{-5mm}
\end{figure}

\noindent \textbf{Performance of Intra-grid Policy. }
The third set of experiments studies the performance of different approaches in generating the Intra-grid policies. Here we use the best performing PPO algorithm to output the inter-grid actions. We compare the proposed ac-PPO with three variants, where we replace the intra-grid policy network $\pi_{\phi}^s$ with rule-based strategies: the random (PPO-RND), the revenue greedy (PPO-REV), and the demand gap greedy (PPO-DMD). Essentially, here we consider a vanilla version of action cascading, where inside the grids we follow certain heuristics to find the destination. For fair comparison, we use the same reward function as in our ac-PPO for all the algorithms. As shown in Table.~\ref{tbl:overall}, the random approach (PPO-RND) produces worse results than the baseline NR. This is expected, as even within the optimal grid, the variations of station can be significant, where selecting a wrong station would hugely affect the reward. This would also have knock-on effects on training the PPO on top, since the obtained rewards can no longer faithfully represent the potential gain of the inter-grid actions. The revenue greedy approach (PPO-REV) is more sensible than PPO-RND, by offering 8\% improvement in demand satisfied rate than the baseline NR. However, as discussed above, this approach tends to perform lots of unnecessary repositions and push EVs to high value sites, causing undesirable performance in net revenue. We observe similar trend in the demand gap greedy algorithm (PPO-DMD), which offers similar demand satisfied rate (about 8\% improvement) and slightly better net revenue (2\% improvement). As expected, the proposed ac-PPO performs the best overall, and the gap between ac-PPO and PPO-DMD is about 10\% in net revenue and 6\% in demand satisfied rate. This confirms that the two sub-actions (inter-grid and intra-grid) should be optimized jointly, and the proposed intra-grid policy network $\pi_{\phi}^s$ outperforms the rule-based baselines under the action cascading framework. 

\begin{figure}[t]
\centering
\begin{subfigure}{0.5\columnwidth}
\includegraphics[width=\textwidth]{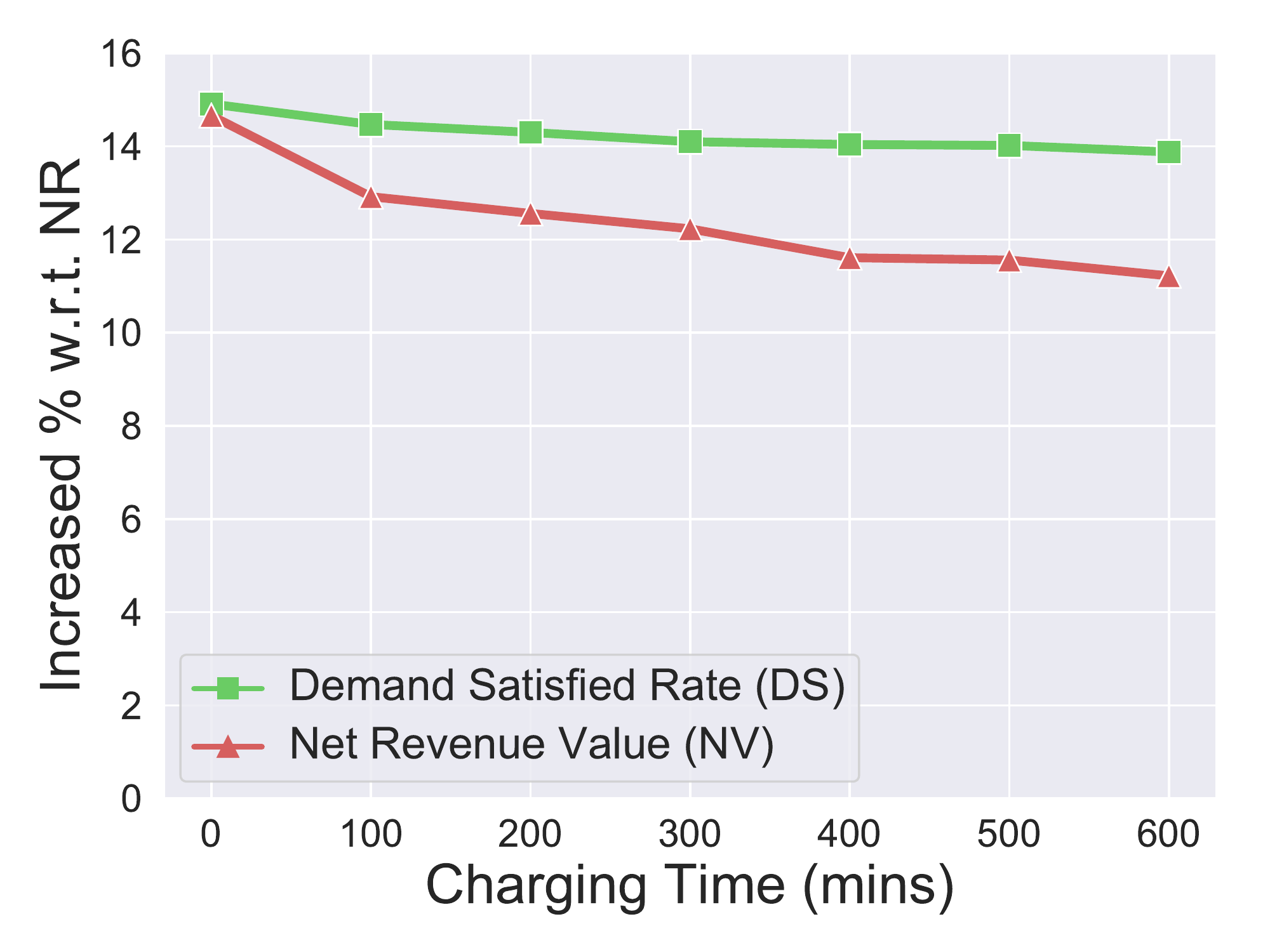}
\caption{ }
\end{subfigure}%~\hspace{5mm}
\begin{subfigure}{0.5\columnwidth}
\includegraphics[width=\textwidth]{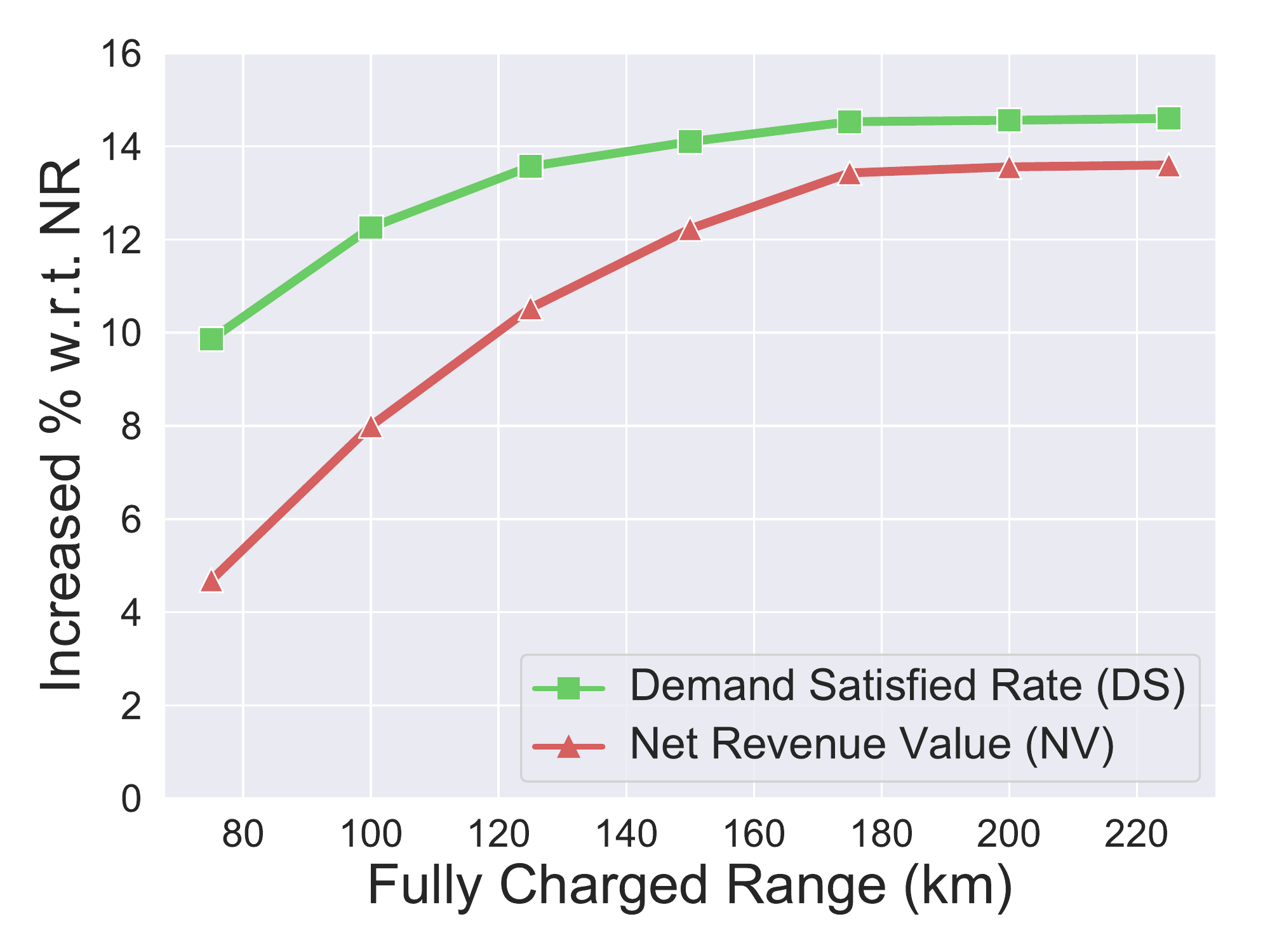}
\caption{}
\end{subfigure}
\caption{Performance of the proposed ac-PPO algorithm vs. (a) charging time, and (b) EV fully charged range.} 
\label{fig:charging}
\vspace{-5mm}
\end{figure}

\noindent \textbf{Impact of System Expansion Dynamics. }
This set of experiments investigate the impact of system expansion dynamics to the rebalancing algorithms. Here we only consider the state-of-the-art STRL and the proposed ac-PPO, as we have shown that the baselines are inferior to both of them in previous experiments. In the experiments, we adjust the simulator to allow different speeds of station network expansion, i.e., on average how many new stations should be deployed and existing stations closed per day. Essentially here we control the level of dynamics in the station network. We vary the speed from 0 to 3, where 0 means the station network is static, and 1 means station network expands at the same speed with that in the real world. As shown in Fig.~\ref{fig:expansion}, we see that when there is no dynamics at all, the gap between STRL and ac-PPO is only about 4\% in demand satisfied rate, and 6\% in net revenue. Also we find that in this case, on average STRL only needs to reposition 1.7 EVs to satisfy an extra order, which is already quite efficient. This is expected, as in this static case STRL clusters the stations into groups, and uses spatial-temporal RL to estimate the future demand of those groups, in order to reposition the EVs accordingly. However, as the system begins to expand, the performance of STRL drops immediately. We already see that at the normal speed, the gap between STRL and our ac-PPO is more than 6\% in demand satisfied rate, and 11\% in net revenue. In the extreme case where the expansion speed is 3$\times$, we see that gap in demand satisfied rate becomes almost 10\%, while STRL can not increase the net revenue when the expansion speed is faster than 1.5. This is also expected, as STRL relies heavily on the station clustering performance, where as the station network is very dynamic, naturally its clustering algorithm would fail to produce optimal results, leading to inferior decisions in rebalancing. On the other hand, we see that the ac-PPO approach is very robust as the expansion speed increases, confirming that the proposed action cascading can work well under different levels of expansion dynamics.

\noindent \textbf{Performance vs. Charging Time. }
In this set of experiments, we study a practical problem in the EV sharing industry: how charging time would affect our rebalancing performance. This is of great importance since one of the key problems of the current EVs is that the the range and charging delays often impact their usage patterns. For instance, the users may behave very differently when driving EVs whose batteries can be replaced immediately, those with super charging, or the ones with normal charging time. In this experiment, we first fix the EV range at 150km when fully charged, and vary the charging time of the EVs from 0 to 600min, where 0 in this case means the batteries of the EVs can be changed instantly. Note that in all other experiments, we assume the charging time of the EVs is 300min, which is consistent with the real data. Fig.~\ref{fig:charging}a show the demand satisfied rate (DS) and net revenue (NV) increased by the proposed ac-PPO algorithm with respect to baseline NR at different charging rate. We see that clearly as charging time increases, the performance gain of our algorithm drops. This is expected because we can not perform any reposition action when the EV is charging. In the extreme case if the EVs are battery replaceable, the increase in NV is about 3\% comparing to the standard case with 300min charging time. This means the approach of replaceable batteries does have its merits in some cases, and should be considered in practice. On the other hand, even in the slowest charging case (600min), our ac-PPO can still improve $>$10\% in NV and about 14\% in DS. The gap between the cases of fastest and slowest charging is negligible in DS, and about 3\% in NV. This means our ac-PPO algorithm is very robust to different charging time: for rebalancing task, even systems with slow chargers could enjoy considerable performance boost. 

\noindent \textbf{Performance vs. Battery Capacity. }
This set of experiments studies the impact of battery capacity i.e. fully charged range of the EVs to the performance of the rebalancing task. This is also a practical problem, which essentially indicates how shared e-mobility systems with different EV models (short range vs. long range) would behave under the proposed algorithm. Here we fix the charging time at 300min and vary the EV range from 75km to 225km. Note that in all the other experiments we use EV range as 150km. Fig.~\ref{fig:charging}b shows the increased demand satisfied rate (DS) and net revenue (NV) of the proposed ac-PPO algorithm compared to the baseline NV. We can see that as the range of the EVs increases, the performance gain becomes more significant. This makes sense because EVs with longer range require less frequently charging, and often allow more flexible rebalancing: they could be repositioned to further stations if needed. We also observe that the performance is more sensitive for EVs with shorter range. For instance, when the range drops from 150km to 75km, the performance gain in NV is halved. However, even in that case our ac-PPO algorithm offers about 10\% improvement in demand satisfied rate, as well as $>$5\% more net revenue value, which is still better than the state-of-the-art. On the other hand, we see that after the range increases over 175km, the extra benefit brought by longer range becomes negligible. This means in practice, EV models with different ranges do react differently to the rebalancing task, but after a certain point the longer range EVs won't contribute much to the performance gain.

\begin{figure}[t]
\centering
\begin{subfigure}{0.48\columnwidth}
\includegraphics[width=\textwidth]{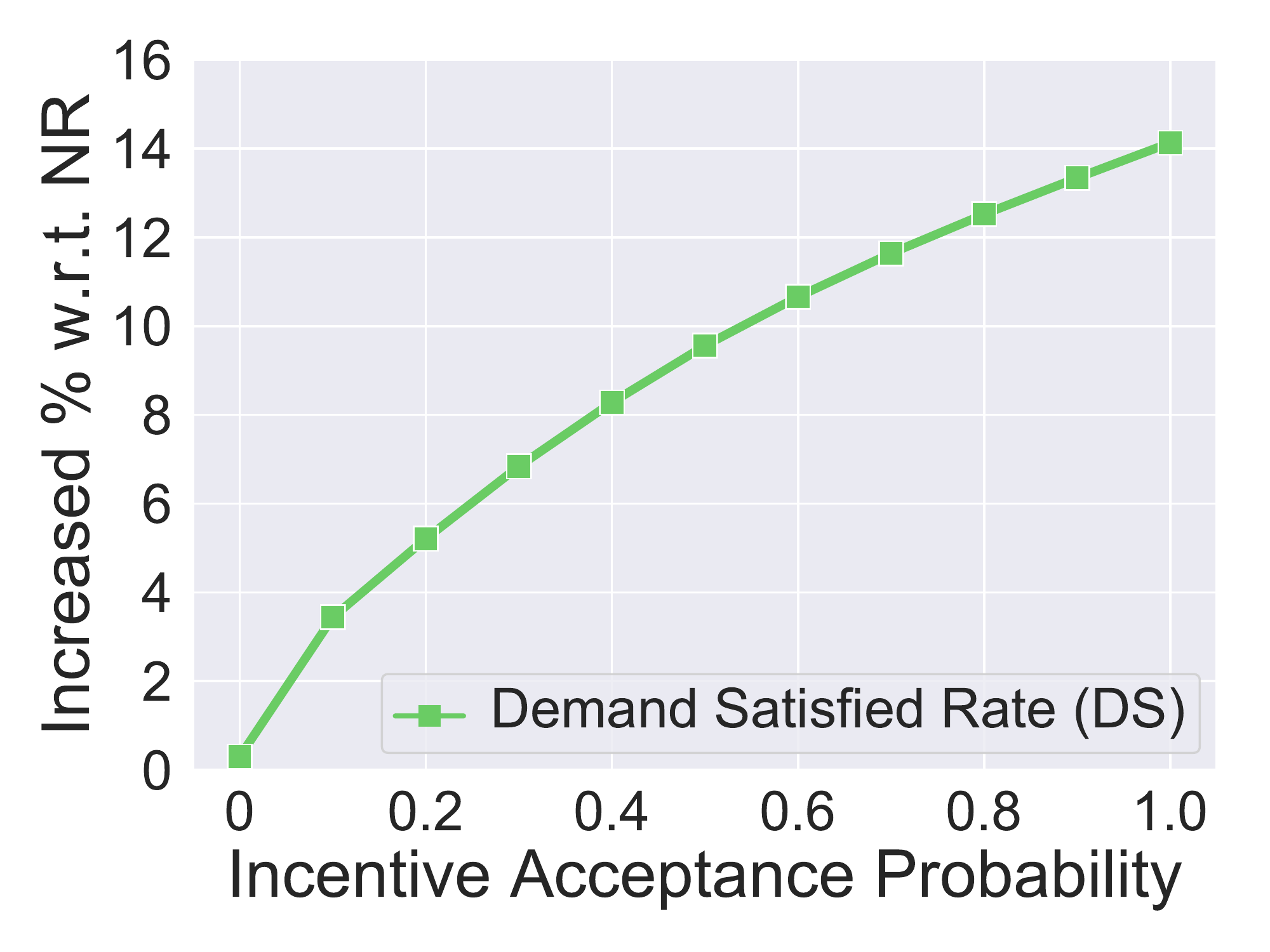}
\caption{ }
\end{subfigure}~\hspace{1mm}
\begin{subfigure}{0.48\columnwidth}
\includegraphics[width=\textwidth]{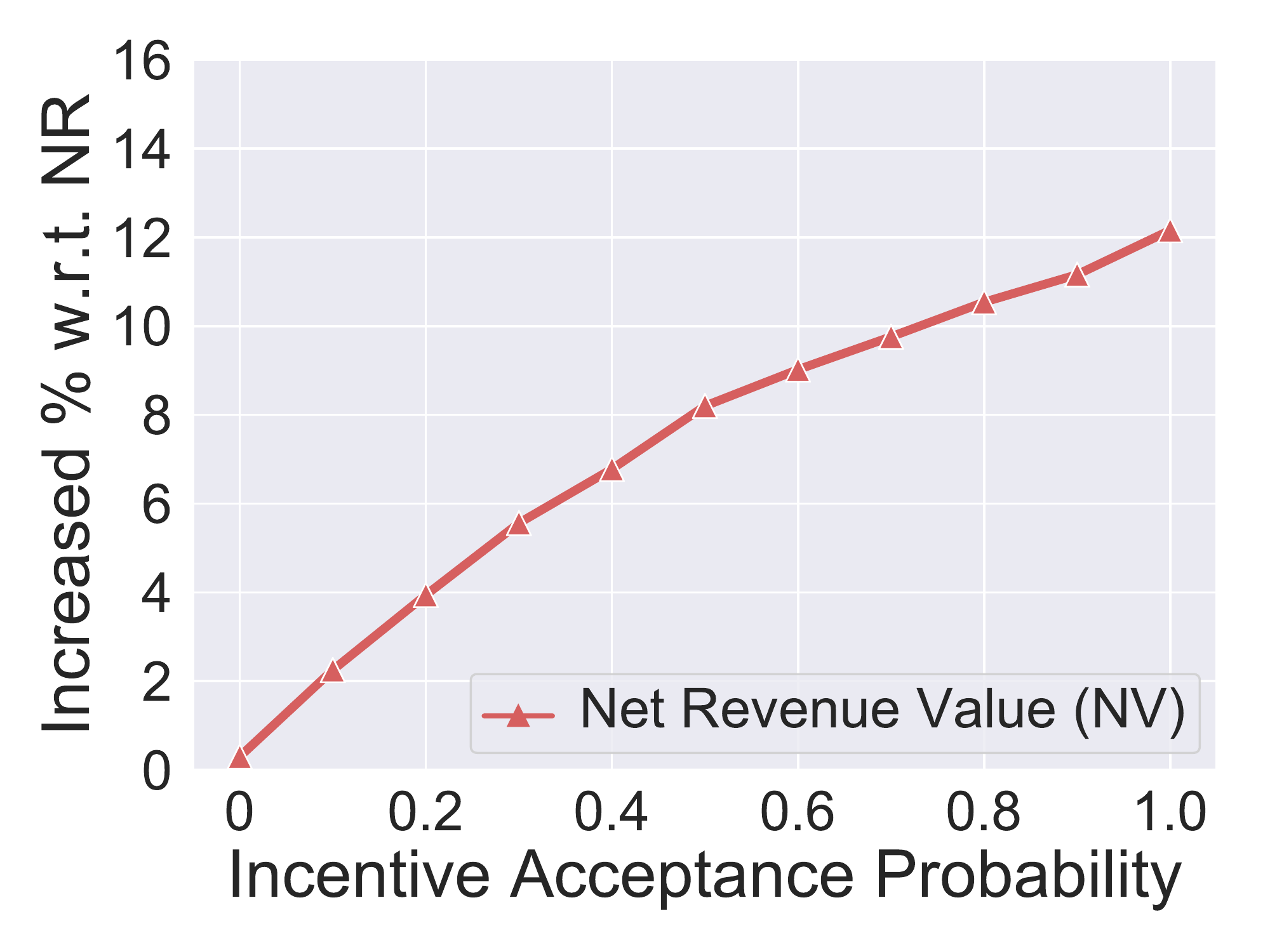}
\caption{}
\end{subfigure}
\caption{Performance of the proposed ac-PPO algorithm vs incentive acceptance probability: (a) increased \% in DS, and (b) increased \% in NV.} 
\label{fig:incentive}
\vspace{-5mm}
\end{figure}

\noindent \textbf{Performance vs. Incentive Acceptance Probability. }
This set of experiments investigates the effectiveness of the proposed ac-PPO algorithm for fleet rebalancing under different user behaviour. In particular, we vary the incentive acceptance rate $p$, as defined in Sec.~{\ref{sub:problem}}, which describes how likely a user would agree to reposition the EVs to alternative stations, i.e., how cooperative they are with respect to the incentive-based rebalancing strategies. Concretely in our simulator, we assume that when presented with a rebalancing offer, a user may accept it with probability $p$. Therefore, by varying $p$ we can obtain different user models, from very cooperative users ($p$ close to 1), to those that are likely to decline any balancing offers (small $p$ towards 0). For simplicity in this paper we assume all users share the same $p$ across space and time. Fig.~{\ref{fig:incentive}} shows the rebalancing performance of our ac-PPO algorithm under different $p$ values. We see that as incentive acceptance probability $p$ increases, both the resulting DS and NV improves. This is expected since as the users become more cooperative, our algorithm can reposition more EVs, and thus smooth the vehicle distribution especially around the busy stations, enabling the system to satisfy more orders in the future. We also observe that as $p$ increases, the gain in both DS and NV slows down, e.g. when $p$ is larger than 0.5. This is because when $p$ is small, i.e. the users tend to reject most of the rebalancing offers, the performance of the system is limited by the amount of rebalancing operations carried out, while on the other hand when the users are very cooperative, the impact of rebalancing becomes saturated. Nevertheless, this also shows that our algorithm degrades gracefully as $p$ decreases, confirming its robustness to different user models. For instance, as we can see in Fig.~{\ref{fig:incentive}}, even with only 50\% of the balancing operations performed (accepted by the users), we can still achieve 10\% improvement in DS, and about 8\% in NV.

\begin{table}[t]
\centering
\small
\begin{tabular}{|@{}c@{}|c|c|c|c|c|c|c|c|c||c|c|c|}
\hline
\multicolumn{1}{|@{}c@{}|}{}  & \multicolumn{1}{@{}c@{}|}{\{2.5e-5;20\}} &  \multicolumn{1}{@{}c@{}|}{\{2.5e-5;40\}} & \multicolumn{1}{@{}c@{}|}{\textbf{\{5e-5;20\}}}   & \multicolumn{1}{@{}c@{}|}{\{10e-5;10\}}  & \multicolumn{1}{@{}c@{}|}{\{10e-5;20\}}   \\ \hline\hline
$\Delta$DS   & 13.88\% & \best{14.14\%} & \textbf{\secondbest{14.10\%}}  & 13.73\% & 12.51\%    \\ \hline
$\Delta$GMV  & 17.86\%      &\best{18.28\%}  &\textbf{\secondbest{18.13\%}}     &17.66\% &15.80\%                      \\ \hline
$\Delta$NV          & 11.78\% & \secondbest{12.13\%} & \textbf{\best{12.23\%}}   & 11.53\% & 10.19\%          \\ \hline
$\Delta |\boldsymbol{o}|$\slash $|\boldsymbol{a}|$           & 1.13 & 1.13 & \textbf{\best{1.12}}  & 1.13 & 1.13           \\ \hline
\end{tabular}
\caption{Performance of the proposed ac-PPO under different \{learning rate $l$; training epochs $e$\} settings. Default settings in bold, best in red and second best in blue.}
\label{tbl:lr}
\end{table}

\begin{table}[t]
\centering
\small
\begin{tabular}{|@{}c@{}|c|c|c|c|c|}
\hline
\multicolumn{1}{|@{}c@{}|}{}  & \multicolumn{1}{c|}{1W} &  \multicolumn{1}{c|}{2W} & \multicolumn{1}{c|}{\textbf{1M}}   & \multicolumn{1}{c|}{3M} & \multicolumn{1}{c|}{STRL(1M)}    \\ \hline\hline
$\Delta$DS   & 12.69\% & 13.30\% & \textbf{\best{14.10\%}}  & \secondbest{14.01}\%  & 7.78\%  \\ \hline
$\Delta$GMV  & 16.15\%      &16.74\%  &\textbf{\best{18.13\%}}     &\secondbest{17.98\%} &  9.27\%                    \\ \hline
$\Delta$NV          & 10.29\% & 10.72\% & \textbf{\best{12.23\%}}   & \secondbest{11.76\%} &    1.12\%      \\ \hline
$\Delta |\boldsymbol{o}|$\slash $|\boldsymbol{a}|$           & 1.13 & 1.13 & \textbf{\best{1.12}}  & 1.13 & 2.08 \\ \hline
\end{tabular}
\caption{Performance of the proposed ac-PPO trained with data of different temporal lengths. Default settings in bold, best performance in red, second best in blue.}
\label{tbl:training-length}
\end{table}

\begin{table}[t]
\centering
\small
\begin{tabular}{|@{}c@{}|c|c|c|c|}
\hline
\multicolumn{1}{|@{}c@{}|}{}  & \multicolumn{1}{c|}{IR} &  \multicolumn{1}{c|}{MR} & \multicolumn{1}{c|}{\textbf{EC}}   & \multicolumn{1}{c|}{STRL (EC)}    \\ \hline\hline
$\Delta$DS   & \secondbest{12.71\%} & 12.65\% & \textbf{\best{14.10\%}}   & 7.78\%  \\ \hline
$\Delta$GMV  & \secondbest{16.18\%}      &16.08\%  &\textbf{\best{18.13\%}}     &  9.27\%                    \\ \hline
$\Delta$NV          & 10.37\% & \secondbest{10.50\%} & \textbf{\best{12.23\%}}   &    1.12\%      \\ \hline
$\Delta |\boldsymbol{o}|$\slash $|\boldsymbol{a}|$           & 1.14 & \secondbest{1.13} & \textbf{\best{1.12}}   & 2.08 \\ \hline
\end{tabular}
\caption{Performance of the proposed ac-PPO trained with data of different spatial coverage. Default settings in bold, best performance in red, second best in blue.}
\label{tbl:training-spatial}
\end{table}

\noindent \textbf{Impact of Learning Rates.}
It is well known that the choice of learning rate plays an important role in all types of deep learning based approaches, which governs the amount that the weights are updated during training. As mentioned in Sec.~{\ref{sub:exp-setting}}, for both the proposed ac-PPO approach and the baselines, we use the default hyper-parameter settings when training the policy networks, which is a standard practice adopted in related literature~\mbox{\cite{li:Kdd:2018,Li:WWW:2019}}. To further study the impact of different learning rates on the proposed ac-PPO algorithm, in this set of experiments we vary the learning rate during training and evaluate the performance of obtained models. We also vary the number of training epochs with different learning rates for fair comparison. In particular, we consider different combinations of \{learning rate $l$; training epochs $e$\}: \{$l$=2.5e-4; $e$=20\}, \{$l$=2.5e-4; $e$=40\}, \{$l$=5e-4; $e$=20\}, \{$l$=10e-4; $e$=10\}, \{$l$=10e-4; $e$=20\}. Note that the default setting used in all other experiments is \{$l$=5e-4; $e$=20\}, and the results are summarized in Table~{\ref{tbl:lr}}. We see that in general, the change of learning rate does impact the overall performance of the algorithm, but only to a limited extent, especially when the training epochs are adjusted accordingly, e.g. double training epochs when learning rate halved. In addition, we find that a smaller learning rate with more training epochs may improve the performance, but only marginally. On the other hand, using a larger learning rate tends to deteriorate the performance, even with more training epochs, as it is well known that with large learning rates training may converge too quickly to suboptimal solutions.

\noindent \textbf{Generalization Capabilities.}
This set of experiments investigates the generalization capabilities of the proposed ac-PPO approach in both temporal and spatial domains when unseen data presents. In particular, we train the proposed ac-PPO with data of different temporal or spatial scales generated by our simulator, and evaluate the performance of those differently trained models on the same testing set. For temporal scales, we consider training data of 1 week (1W), 2 weeks(2W), 1 month (1M) and 3 months (3M), while for spatial scales we consider data generated within the inner ring road of the city (denoted as IR), the middle ring road (MR), and across the entire city (EC). Note that the default settings used in other experiments are 1M and EC. Table~{\ref{tbl:training-length}} and Table~{\ref{tbl:training-spatial}} show the performance of the different variants. We see that in general training with more data offers better performance, while our approach demonstrates strong generalization capabilities, e.g., when trained with much less data, the performance drop compared to the default setting is typically less than 2\% across all metrics. Even models trained with very limited data (1W or IR) significantly outperform the baseline STRL~{\cite{li:Kdd:2018}}.

%%%%%%%%%%%%%%%%%%%%%%%%%%%%%%%%%%%%%%%%%%%%%%%%%%%
%% Related
%%%%%%%%%%%%%%%%%%%%%%%%%%%%%%%%%%%%%%%%%%%%%%%%%%%

\section{Related Work}
\label{sec:related}

\noindent \textbf{Shared Mobility Systems. } %range
Recently, shared mobility systems have attracted extensive interest from different research communities interest~\cite{jiang:www:2018, xu:kdd:2018, furuhata:transportation:2013, li:ICIS:2016, dillahunt:HCI:2017, kooti:WWW:2017}. Shared E-mobility systems, although relatively new, have also exposed various new problems and challenges, such as route planning and optimization~\cite{Sarker:IMWUT:2018, yuen:WWW:2019}, charging scheduling~\cite{Yan:IMWUT:2018,Yuan:ICDCS:2019,Wang:IMWUT:2019}, and infrastructure planning~\cite{Sarker:IMWUT:2018, du:kdd:2018}. The work in~\cite{Wang:MobiCom:2019} conducted a comprehensive measurement investigation to study the long-term evolving mobility and charging patterns of electric taxis in a city, using real-world data collected over five years. Our work complements existing studies which primarily consider the electric taxis or buses, in that we focus on the EV sharing systems, which operate in a very different way. For instance, for electric taxis or buses their networks of stations~\cite{Wang:MobiCom:2019} are mainly used for charging, whose service coverage also depends on the fixed bus routes or individual taxi drivers. However in our case, the users can only access (renting, returning and charging) vehicles at the available stations, and thus the expansion of the station network will have much more direct and complex impact on the entire system. Moreover, comparing to traditional vehicle sharing systems, EV sharing systems are more complex due to the unique properties of EVs. For example, the existing works focused on rebalancing shared mobility system~\cite{li:Kdd:2018, singla:AAAI:2015, lin:Kdd:2018, Liu:KDD:2016} do not need to worry about the remaining range of their vehicles, nor the charging time. 

\noindent \textbf{Fleet Rebalancing in Shared Mobility. } %dynamic station network
Existing work to address the problem of rebalancing the shared mobility services can be broadly categorized into three types, static reposition, dynamic reposition and user-based reposition. The first two are conducted by the system operators while the last is conducted by users. The static reposition solution is usually performed when the system is not operating or during the nights, with no perturbation from the users. Then the rebalancing taks is cast into an optimization problem with some objectives~\cite{Liu:KDD:2016, raviv:EUROJ:2013,zhang:kdd:2017}, e.g., maximizing satisfaction of the customers. In practice, such static repositioning approaches only work well if the demand is predictable and stable. However, they can't perform rebalancing online as during operation the distributions of vehicles are varying. Dynamic reposition approaches consider the real-time flow of vehicles in the system, which also use optimization techniques~\cite{ghosh:IJCAI:2016, singla:AAAI:2015, ghosh:AIJ:2017, li:Kdd:2018, wang:WWW:2019, etienne:TIST:2014} to find the optimal repositioning plans. However, they depend heavily on the accuracy of demand prediction and it is often difficult to adjust the reposition operation given the unpredictable fluctuations in demand. The user-based reposition approaches solve the problem by incentivizing the users with rewards to rent or return vehicles at specific stations~\cite{singla:AAAI:2015, pan:AAAI:2019, chemla:EUROJ:2013}. However, it is often challenging to determine the optimal alternative station and estimate the appropriate reward to offer. This paper falls into the last category, but unlike the existing solutions which assumes the system is static, we aim to tackle the rebalancing problem in the presence of dynamically changing station networks. It is fundamentally different from the static case, as at different time the candidate stations for potential repositioning operations can be different, which can't be addressed by the existing approaches. 

\noindent \textbf{Deep Reinforcement Learning in Mobility. } % rl
Deep reinforcement learning techniques have been used in various ubiquitous mobility applications due to their superior performance, such as traffic management~\cite{wei:IEEEACESS:2017look, bakker:ICIS:2010}, order dispatching~\cite{Li:WWW:2019,lin:Kdd:2018}, and rebalancing~\cite{li:Kdd:2018,pan:AAAI:2019}. Due to their distributed nature, many of those mobility applications can be modeled as multi-agent games, which can be well solved by deep reinforcement learning. For instance, the work in~\cite{Li:WWW:2019} addresses the order dispatching problem for ride sharing systems using mean field multi-agent reinforcement learning, while~\cite{lin:Kdd:2018} proposes a contextual multi-agent reinforcement learning framework to tackle the fleet management problem. Another work in~\cite{li:Kdd:2018} considers a spatio-temporal reinforcement learning approach, to dynamically reposition bikes across different areas in the bike-sharing system. It has been shown that in those applications, deep reinforcement learning often achieves better performance, e.g. in terms of reducing potential customer loss, or increasing the gross merchandise values, than the traditional rule-based or optimization based approaches, especially when the problem structure is very complicated. In this paper, we also model the rebalancing problem in EV sharing system with the framework multi-agent reinforcement learning. However, our work differs from the existing work in that a) we only consider the reposition of the EVs but not the pricing strategies for user incentives, b) we directly model the unique properties of EV sharing such as range limitations and charging time into the learning framework, and finally c) we extend the existing RL algorithms to support dynamically changing action spaces, i.e. the candidate stations for repositioning, with the proposed action cascading techniques.

%%%%%%%%%%%%%%%%%%%%%%%%%%%%%%%%%%%%%%%%%%%%%%%%%%%
%% Conclusion
%%%%%%%%%%%%%%%%%%%%%%%%%%%%%%%%%%%%%%%%%%%%%%%%%%%

\section{Conclusion}
In this paper, we investigate fleet rebalancing for expanding shared e-mobility systems by incentivizing users. We conduct a thorough analysis of a real-world shared e-mobility system with data collected over a year, and design a simulator to capture its operational details in practice. Based on the gained insights, we formulate the rebalancing task as a Multi-agent Reinforcement Learning problem, and tackle it with a novel policy optimization approach with action cascading to handle the non-stationarity induced by dynamic expansion. Extensive experiments have shown that our approach significantly outperforms the baselines and state-of-the-art in both satisfied demand and revenue, and is robust to different system expansion dynamics, EV charging time/range, and user models. Despite the superior performance, the proposed approach also has its limitations. For instance, it is designed as a data-driven paradigm and thus requires substantial data to calibrate the simulator and train the deep reinforcement learning algorithms, which may not be possible when data access is constrained. In addition, when applying to new context, e.g. another city, our approach may need re-training. Finally, the current reward function only pushes towards demand satisfaction and net revenue, while fairness may become an issue in certain cases. For future work, we would like to improve the practicality and generalization capability of our approach, e.g. combining rule-based rebalancing with our learning-based approach for better flexibility, incorporating techniques such as transfer learning to make it work well in new cities, considering more sophisticated user models to capture the realistic user behaviour, supporting more types of rebalancing operations, e.g. for both pick-up and return, and improving the fairness of the rebalancing strategies.

\smallskip
\smallskip
\noindent \textbf{Acknowledgment} 
This work was supported in part by The Alan Turing Institute under the EPSRC grant EP/N510129/1.

\bibliographystyle{IEEEtran}
\bibliography{10-main}

\end{document}